\documentclass[10pt,twocolumn,letterpaper]{article}

\usepackage{cvpr}              

\usepackage{graphicx}
\usepackage{amsmath}
\usepackage{amssymb}
\usepackage{booktabs}
\usepackage{verbatim}
\usepackage{multirow}
\usepackage{makecell}
\usepackage{enumitem}
\usepackage[svgnames]{xcolor}
\definecolor{lightblue}{RGB}{46,150,222}

\newcommand{\tx}[1]{\textcolor[rgb]{0,.0,0}{#1}}
\newcommand{\rynq}[1]{\textcolor{black}{#1}}
\newcommand{\kk}[1]{\textcolor[rgb]{0,.0,0}{#1}}
\newcommand{\ttxx}[1]{\textcolor[rgb]{0,.0,0}{#1}}

\usepackage[pagebackref,breaklinks,colorlinks]{hyperref}
\hypersetup{colorlinks,breaklinks,
             citecolor=lightblue}

\usepackage[capitalize]{cleveref}
\crefname{section}{Sec.}{Secs.}
\Crefname{section}{Section}{Sections}
\Crefname{table}{Table}{Tables}
\crefname{table}{Tab.}{Tabs.}

\def\cvprPaperID{7870} 
\def\confName{CVPR}
\def\confYear{2022}

\begin{document}

\title{Bi-directional Object-Context Prioritization Learning for Saliency Ranking}

\author{\large{Xin Tian$^{1,2,3}$\quad Ke Xu$^{2,\dagger}$\quad Xin Yang$^{1,\dagger}$\quad Lin Du$^{3}$\quad  Baocai Yin$^{1}$\quad Rynson W.H. Lau$^{2,\ddagger}$}\\
\normalsize{$^{1}$Dalian University of Technology\quad  $^{2}$City University of Hong Kong} \\ \normalsize{$^{3}$AI Application Research Center (AARC), Huawei Technologies Co., Ltd.}
\\
\tt\small$\lbrace$xin.tian.831, kkangwing$\rbrace$@gmail.com, $\lbrace$xinyang, ybc$\rbrace$@dlut.edu.cn, \\\tt\small{dulin09@huawei.com, Rynson.Lau@cityu.edu.hk}
}
\maketitle

\renewcommand{\thefootnote}{}
\footnotetext{\textsuperscript{$\dagger$} Ke Xu and Xin Yang are joint corresponding authors.}
\footnotetext{\textsuperscript{$\ddagger$} Rynson Lau leads this project.}

\begin{abstract}

The saliency ranking task is recently proposed to study the visual behavior that humans would typically shift their attention over different objects of a scene based on their degrees of saliency.
Existing approaches focus on learning either object-object or object-scene relations. Such a strategy follows the idea of \textbf{object-based attention} in Psychology, but it tends to favor objects with strong semantics (\eg, humans),
resulting in unrealistic saliency ranking.
We observe that \textbf{spatial attention} works concurrently with {object-based attention} in the human visual recognition system.
During the recognition process, the human 
\kk{spatial attention mechanism would move, engage, and disengage from region to region (\ie, context to context). This inspires us to model region-level interactions, in addition to object-level reasoning, for saliency ranking.
Hence,} we propose a novel bi-directional method to unify spatial attention and object-based attention for saliency ranking.
Our model has two novel modules:
(1) a selective object saliency (SOS) module to model object-based attention via inferring the semantic representation of salient objects,
and (2) an object-context-object relation (OCOR) module to allocate saliency ranks to objects by jointly modeling object-context and context-object interactions of salient objects.
Extensive experiments show that our approach outperforms existing state-of-the-art methods.
\ttxx{Code and pretrained model are available at \url{https://github.com/GrassBro/OCOR}.}

\vspace{-4mm}
\end{abstract}

\section{Introduction}
\label{sec:intro}

Saliency detection is a fundamental task in computer vision.
Previous works mainly focus on the salient object / instance detection tasks~\cite{fan2019shifting,he2015supercnn,tian2020weakly}, and the gaze prediction task~\cite{recasens2017following}. 
Recently, Siris \etal~\cite{siris2020inferring} propose to study a novel task called saliency ranking, which aims to detect the salient instances and infer their saliency ranks simultaneously. 
By mimicking \ttxx{how humans change their attention} across the scenes \ttxx{depending on} the saliency ranks,
saliency ranking can benefit many down-stream visual tasks, \eg, image manipulation~\cite{chen2016automatic,NEURIPS2018active}, scene understanding~\cite{9591338}, \ttxx{important person identification~\cite{li2019learning} and their interaction reasoning~\cite{fan2019understanding}}.

\fboxsep=0.15mm
\fboxrule=0.15mm
\begin{figure}[!t]
\includegraphics[width=.45\textwidth]{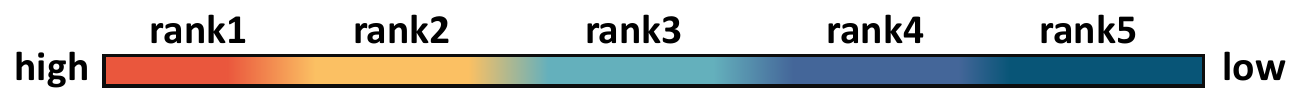}
\vspace{0.14cm}
\\
\centering
\includegraphics[width=.075\textwidth]{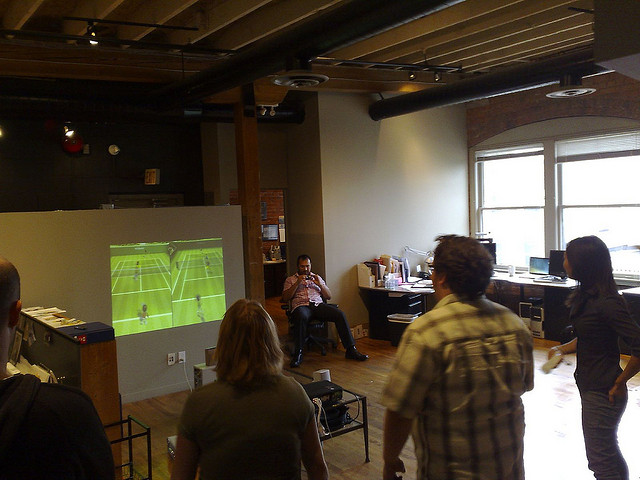}
\centering
\includegraphics[width=.075\textwidth]{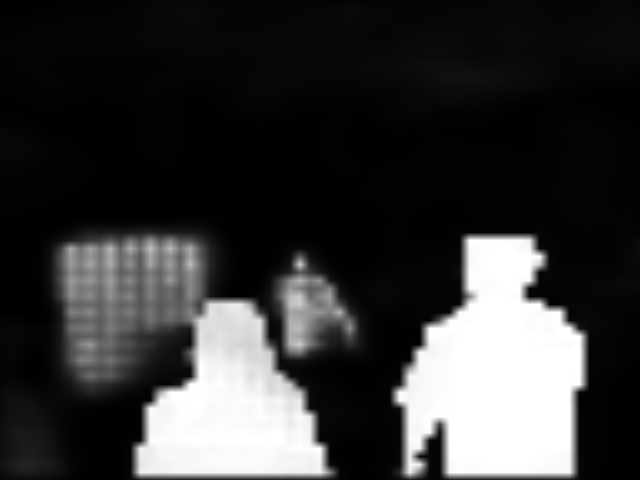}
\centering
\includegraphics[width=.075\textwidth]{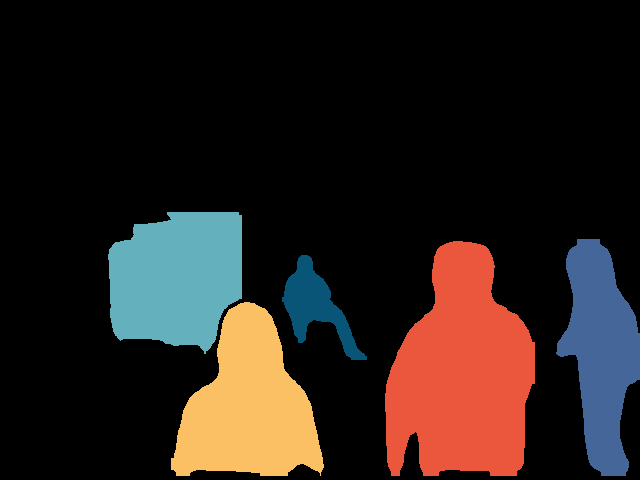}
\centering
\includegraphics[width=.075\textwidth]{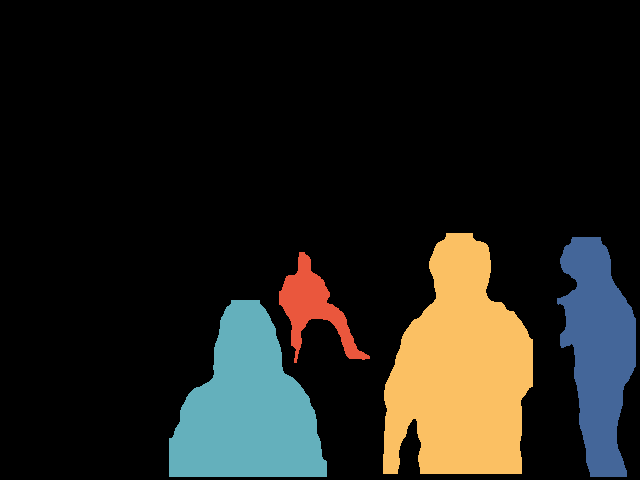}
\centering
\includegraphics[width=.075\textwidth]{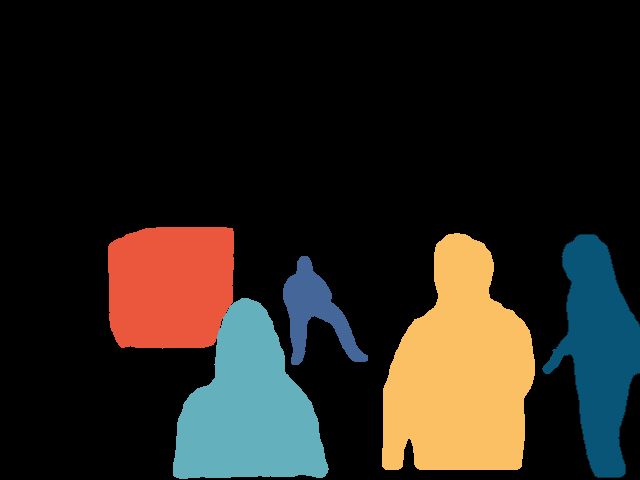}
\centering
\includegraphics[width=.075\textwidth]{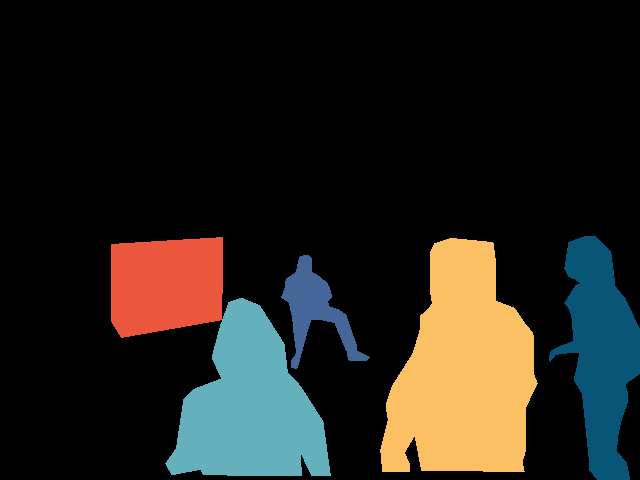}
\\
\centering
\includegraphics[width=.075\textwidth]{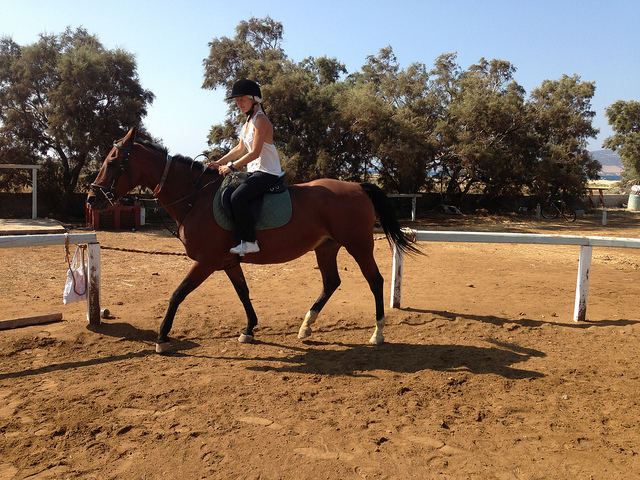}
\centering
\includegraphics[width=.075\textwidth]{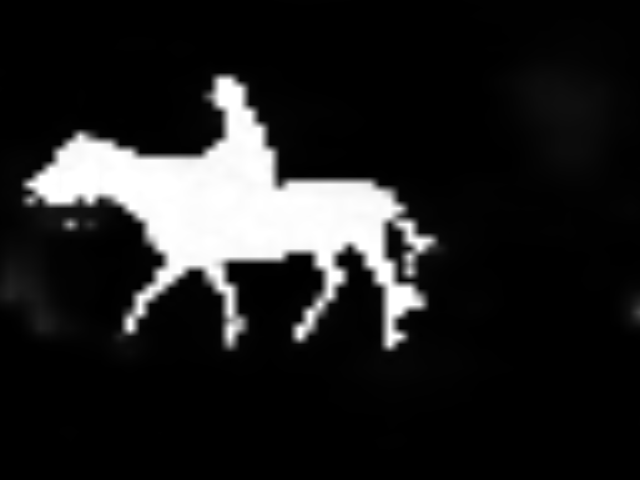}
\centering
\includegraphics[width=.075\textwidth]{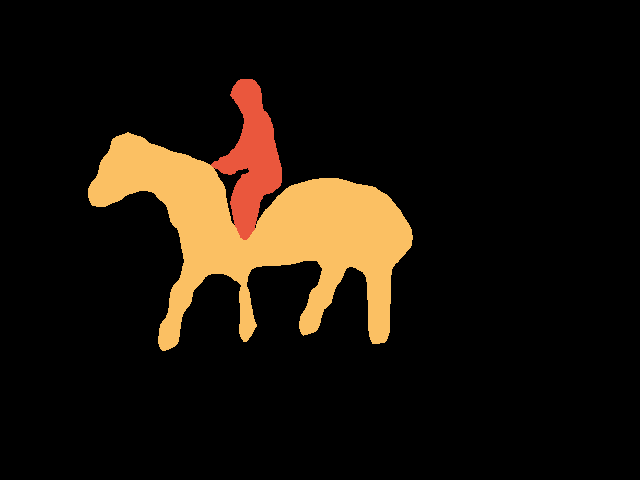}
\centering
\includegraphics[width=.075\textwidth]{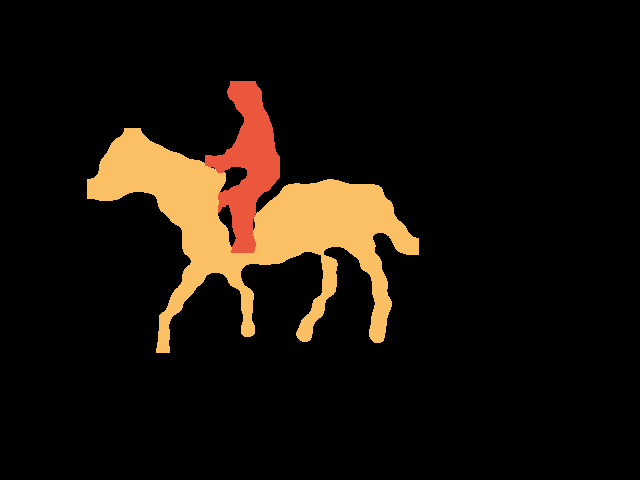}
\centering
\includegraphics[width=.075\textwidth]{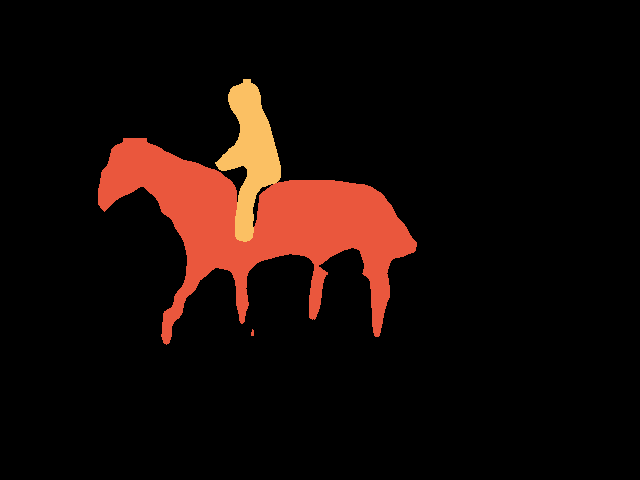}
\centering
\includegraphics[width=.075\textwidth]{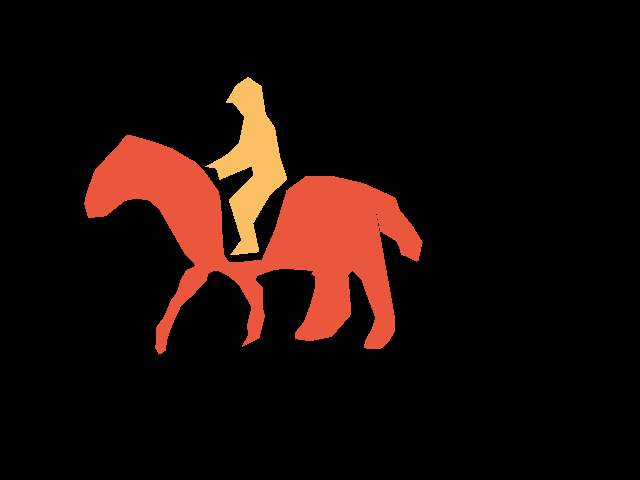}
\\
\centering
\includegraphics[width=.075\textwidth]{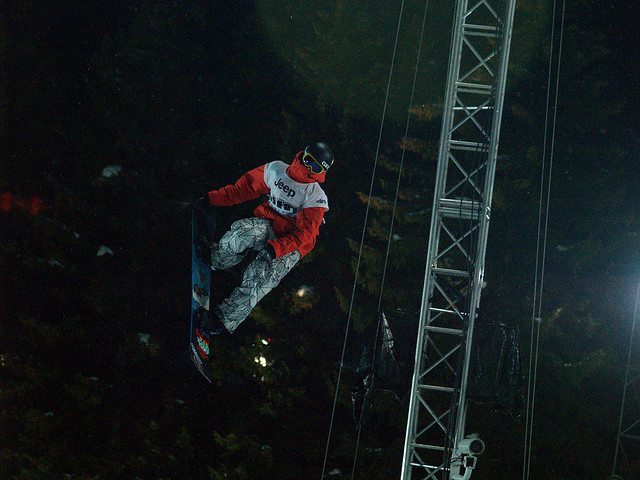}
\centering
\includegraphics[width=.075\textwidth]{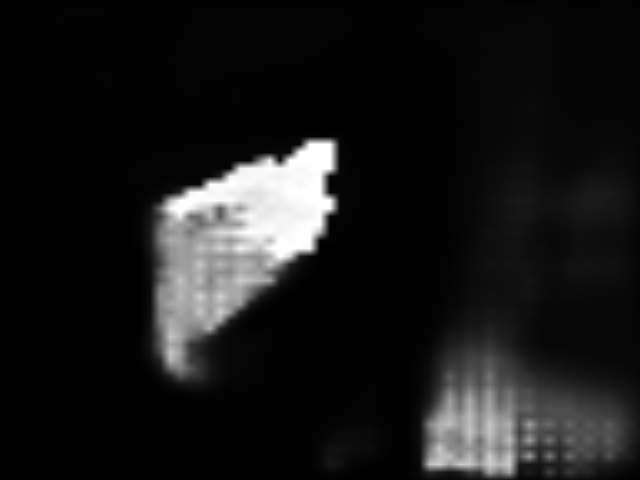}
\centering
\includegraphics[width=.075\textwidth]{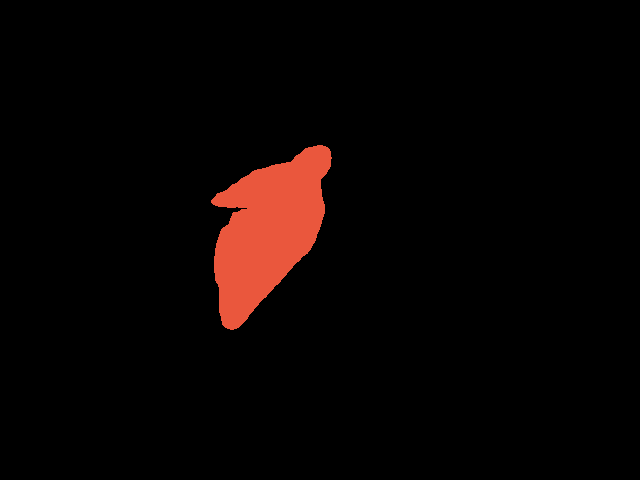}
\centering
\includegraphics[width=.075\textwidth]{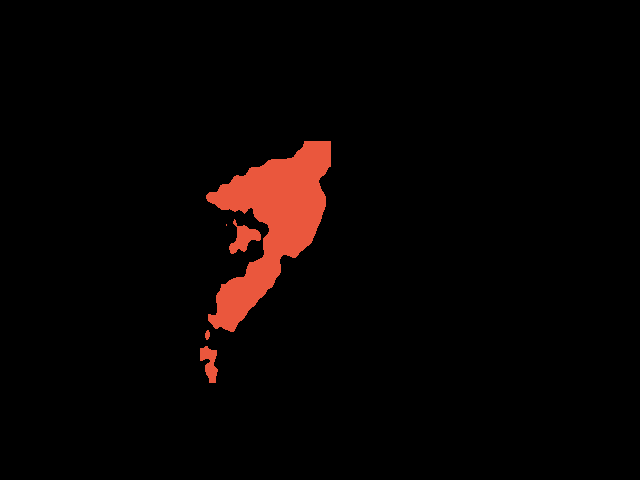}
\centering
\includegraphics[width=.075\textwidth]{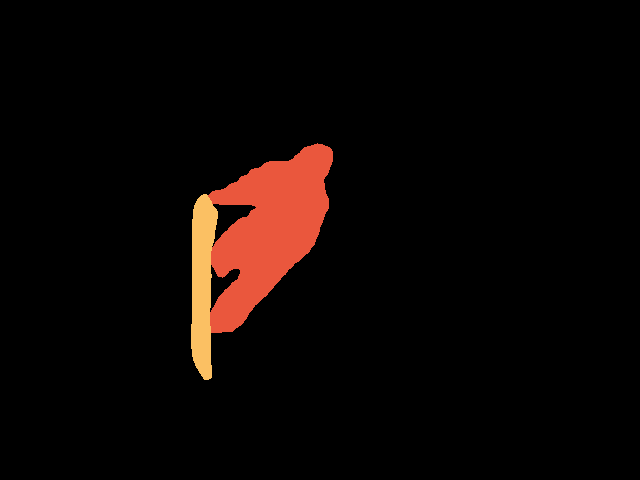}
\centering
\includegraphics[width=.075\textwidth]{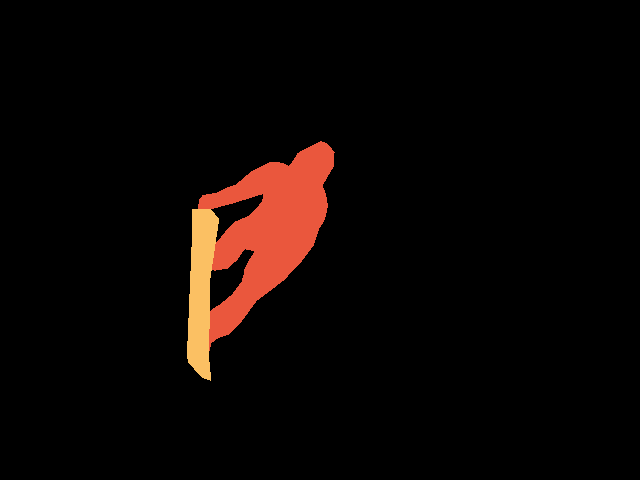}
\\
\centering
\begin{minipage}[t]{0.075\textwidth}
\centering
{{\scriptsize{(a) Image}}}
\end{minipage}
\begin{minipage}[t]{0.075\textwidth}
\centering
{{\scriptsize{(b) RSDNet\\ \vspace{-0.15cm} \cite{islam2018revisiting}}}}
\end{minipage}
\begin{minipage}[t]{0.075\textwidth}
\centering
{{\scriptsize{(c) ASSR\\ \vspace{-0.15cm} \cite{siris2020inferring}}}}
\end{minipage}
\begin{minipage}[t]{0.075\textwidth}
\centering
{{\scriptsize{(d) IRSR \\ \vspace{-0.15cm} \cite{liu2021instance}}}}
\end{minipage}
\begin{minipage}[t]{0.075\textwidth}
\centering
{{\scriptsize{(e) Ours}}}
\end{minipage}
\begin{minipage}[t]{0.075\textwidth}
\centering
{{\scriptsize{(f) GT}}}
\end{minipage}
\vspace{-0.22cm}
\caption{Existing salient ranking approaches~\cite{islam2018revisiting,siris2020inferring,liu2021instance} produce unrealistic saliency \ttxx{ranks}. (b) RSDNet~\cite{islam2018revisiting} is a pixel-level method and does not predict object-level saliency ranks well. (c) ASSR~\cite{siris2020inferring} and (d) IRSR~\cite{liu2021instance} explore object-object and object-scene relations for inferring saliency ranks. However, they favor objects with strong semantics and \kk{tend to assign people with high saliency ranks}. (e) Our method explores both spatial and object-based attentions with a bi-directional object-context prioritization learning formulation, yielding faithful saliency ranking results.
}
\label{fig:teaser}
\vspace{-4mm}
\end{figure}

 Islam~\etal\cite{islam2018revisiting} propose the first saliency ranking work, which directly predicts a relative saliency map with different pixel values indicating different saliency degrees, as shown in Figure~\ref{fig:teaser}(b). However, this method only studies the relative saliency of pixels.
%
Later, Siris~\etal\cite{siris2020inferring} propose to study the salient object ranking as humans shift \ttxx{attention} from object to object.
They propose to model the relations between  objects and global context for reasoning their ranks.
%
%
Liu~\etal~\cite{liu2021instance} further propose a neural graph-based method to learn relations between objects and local contexts as well as relations between objects.
%
%
%
\kk{However, these two methods~\cite{siris2020inferring,liu2021instance} tend to favor objects with \rynq{strong semantics} (\eg, people) as shown in Figure~\ref{fig:teaser}, resulting in incorrect saliency \ttxx{ranks}.
%
For example, in the first and second rows of Figure~\ref{fig:teaser}(c,d), they assign people with highest saliency ranks, despite the \ttxx{visual} distinctiveness of the green screen in the first example and the relatively larger horse in the second example. 
In the last row, the two methods do not even consider the skateboard as a salient object.}

In this paper, we tackle the saliency ranking problem based on the observation that as revealed by psychological studies~\cite{egly1994shifting,anderson2004integrated}, \textbf{spatial attention} and \textbf{object-based attention} work concurrently in the human visual system. 
\kk{While object-based attention directs our views to the candidate objects or perceptual groups via a preattentive segmentation of the scene~\cite{egly1994shifting},
the human spatial attention mechanism allows us to process the scene sequentially through prioritizing the regions (where the objects belong to) based on low-level visual stimuli (\eg, rich colors), functionalities of the objects, and interactions among the objects.}
This inspires us to jointly exploit spatial attention and object-based attention for saliency ranking.
%
%
Based on this observation, we formulate a bi-directional object-context prioritization learning approach to model both region-level and object-level relations.
%
%
We first propose a Selective Object Saliency (SOS) module to model object-based attention via inferring and enriching the semantic representations of the salient objects based on their local contexts.
\kk{We then propose an object-context-object relation (OCOR) module to exploit the spatial attention mechanism by reasoning in the object-context and context-object bi-directional way.} We formulate a multi-head attention mechanism to model the way that an object with its context would interact with other objects with their contexts.
%
\kk{As shown in Figure~\ref{fig:teaser}(e), our approach produces more faithful saliency \ttxx{ranks} over the state-of-the-art methods.
For example, our method can detect the screen (first row) and the horse (second row) as the most salient objects according to their visual distinctiveness. 
In the third row, our method can detect and rank the skateboard by modeling its interaction with the person.
}
%
%
%


To summarize, this work has three main contributions: 
\emph{\textbf{1)}} Inspired by the psychological studies, we propose a novel bi-directional object-context prioritization learning approach for saliency ranking, by jointly exploiting spatial and object-based attention mechanisms.
\emph{\textbf{2)}} We propose a novel selective object saliency (SOS) module for modeling object-based attention, and a novel object-context-object relation (OCOR) module for modeling spatial attention via inferring the relations of objects in a bi-directional object-context and context-object manner.
\emph{\textbf{3)}} We conduct extensive experiments to analyze our approach and verify its superior performance over the state-of-the-art methods.
\vspace{-2mm}

\section{Related Work}
\subsection{Saliency Ranking}
Saliency ranking is a new task. It studies the visual phenomenon \ttxx{that objects in a daily} scene generally have different saliency degrees, which \ttxx{draws} the observer's attention sequentially across the objects.
Islam \etal~\cite{islam2018revisiting} make an initial attempt on this problem, but they only study the pixel-level relative saliency.
%
\ttxx{This is evident from their data collection step,} where multiple objects are sometimes annotated with the same saliency rank.
Siris \etal~\cite{siris2020inferring} propose a new dataset and a model to facilitate research on this task.
Specifically, they exploit the statistics of fixations on objects to build a large-scale dataset, and design a network with an object-context relation module to learn saliency ranks.
Liu~\etal~\cite{liu2021instance} further propose another dataset with less annotation noise, and investigate object-object relations for the task.
Similarly, Fang~\etal\cite{fang2021salient} also propose to model object-object relations but embed the spatial coordinates of objects as spatial cues in the learning step.

By modeling object-based attention, the above methods can take advantages of the learned semantic representations of objects. However, they tend to produce unreasonable saliency \ttxx{ranks}, \eg, they tend to rank humans higher than stuff with \ttxx{visually distinguishing hues}.
In contrast, in this paper, we propose a bi-directional object-context prioritization method to combine spatial attention with object-based attention for saliency ranking.

\subsection{Salient Object Detection (SOD)}
SOD is a long-standing problem with a lot of methods proposed. It aims to detect conspicuous objects in a scene. 
%
Earlier methods~\cite{yang2013saliency,achanta2009frequency,perazzi2012saliency,cheng2014global} mainly rely on hand-crafted features (\eg, color, brightness, and textures) to detect salient objects.
These methods often fail in complex scenarios due to the limited representation capacities of low-level hand-crafted features.
Recent deep learning-based methods \cite{wang2021salient} have achieved superior performances. They mainly incorporate two kinds of deep techniques, {\it deep feature fusion}~\cite{Wei_2020_CVPR,zhang2018bi} and {\it feature attention}~\cite{zhang2018progressive,liu2018picanet,zhao2019pyramid,siris2021scene}. 
Deep feature fusion aims to aggregate multi-level context information including low-level stimulus and high-level semantics for SOD, while feature attention reweights multi-scale features and enhances the context learning to help the model focus on the salient regions and suppress noise from the background regions.

Unlike SOD, the saliency ranking task needs to detect salient instances and then determine their saliency ranks. Hence, existing salient object detection methods cannot be directly applied to the saliency ranking task.


\begin{figure*}[!t]
    \centering
    \includegraphics[width=0.98\textwidth]{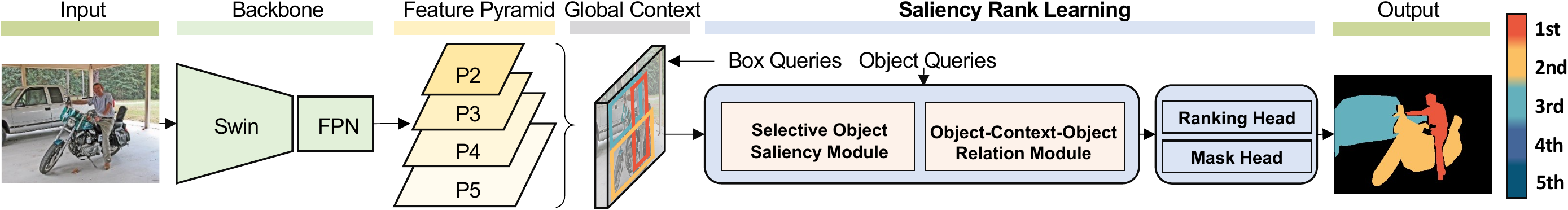}
    \vspace{-2mm}
    \caption{Overview of our proposed network. Given an input image, we first apply the query based object detection method~\cite{sun2021sparse,fang2021instances} to extract global context features and leverage a set of learnable salient object proposals (\ie, box and object queries that encode object locations and rich object characteristics),
    \tx{to help predict the \ttxx{final} saliency \ttxx{ranks}. The Saliency Rank Learning process has (1) a SOS module to capture and enrich object-level semantic representation, (2) \ttxx{an} OCOR module to model the interactions from one object with its context to other objects with their contexts, \ie, bi-directional object-context and context-object relations learning, and (3) ranking and mask heads to infer object-wise saliency ranks on top of the improved features from the SOS and OCOR modules.} 
    }
    \vspace{-4mm}
    \label{fig:pipeline}
\end{figure*}

\subsection{Salient Instance Detection (SID)}
A few methods are proposed to detect salient objects \ttxx{at the instance level.}
Zhang~\etal~\cite{zhang2016unconstrained} propose a Maximum a Posteriori optimization based method to detect salient instances with bounding boxes.
Li~\etal~\cite{li2017instance} propose to leverage the instance-aware saliency contours for detecting instance-level objects.
Fan~\etal~\cite{fan2019s4net} propose to combine the object detection model FPN~\cite{lin2017feature} with a segmentation branch to detect salient instances.
Tian~\etal~\cite{tian2020weakly,tian2022learning} propose a weakly-supervised method that exploits class labels and subitizing labels for SID.

SID methods can provide instance-level information of salient objects, but they do not attempt to rank the saliency degrees of the detected salient instances.

\section{Methdology}
Psychological studies \cite{anderson2004integrated,egly1994shifting} reveal that {\bf spatial attention} and {\bf object-based attention} work cooperatively in the human visual systems in order to process multiple visual inputs in a scene sequentially.
On the one hand, object-based attention tends to fix humans' views to the candidate objects by preattentive segmentation of the scene, as humans are more likely to be attracted to objects that they are most familiar with. On the other hand, spatial attention directs humans' views from region to region via a joint process of multiple factors, such as regional visual stimuli, functionalities of objects, and their contextual interactions.
%
%
This motivates us to exploit object-based attention and spatial attention to design our saliency ranking model.

Figure~\ref{fig:pipeline} shows the overview of our bi-directional object-context prioritization model.
Given an input image, we first use the query-based object detection method~\cite{sun2021sparse,fang2021instances} to extract global features and generate a set of \tx{object features based on} object proposals, \ie, box and object queries that encode object locations and rich object characteristics.
\tx{We then feed them into two novel modules, Selective Object Saliency (SOS) module and Object-Context-Object Relation (OCOR) module, which model the two attention mechanisms for reasoning the saliency ranks.}
Finally, our model learns prioritization information of objects for saliency ranking.

\subsection{Selective Object Saliency Module}
\begin{figure}[!t]
    \centering
    \includegraphics[width=.45\textwidth]{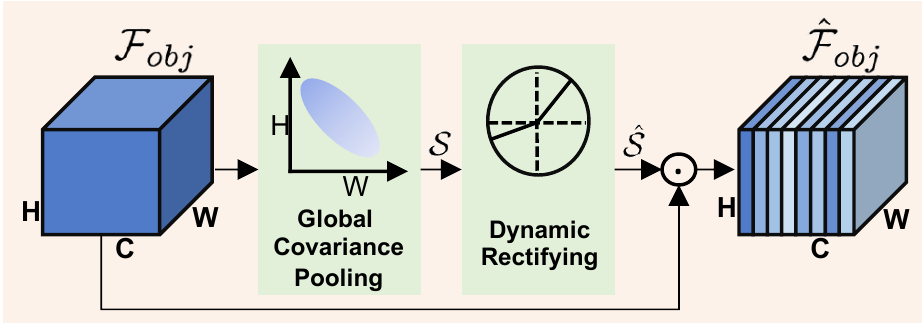}
    \vspace{-2mm}
    \caption{\ttxx{Structure of our SOS module.}}
    \label{fig:sos}
    \vspace{-4mm}
\end{figure}

Following the spirit of the object-based attention mechanism, the goal of our SOS module is to capture and enhance the semantic representation of salient objects.
It has been shown that channels of deep features response to diverse semantic components~\cite{chen2021crossvit,zhou2016learning}.
Previous channel-wise attentions~\cite{woo2018cbam,hu2018squeeze} mainly aim to highlight the discriminative channels according to the ground truth categories while suppressing the responses from other channels.
%
However, simply suppressing the low responses from the other channels may not be suitable for saliency ranking, as these less discriminative channels may also be informative. They may serve as the contexts to correlate objects to each other or as the global context.
Hence, we propose to extend existing channel-wise attentions in two aspects. First, we leverage the global covariance pooling~\cite{wang2020deep,li2017second} to learn object representations as well as their correlations to both local and global contexts. Second, we learn a group of dynamic rectifying functions to reallocate attentions to the channels based on the high-order feature statistics computed by the global covariance pooling. \ttxx{Consequently, they jointly capture fine-grained object information for learning object-based representations.}
%
Figure~\ref{fig:sos} shows our SOS module.

\vspace{1mm}
\noindent{\bf Global Covariance Pooling}
\ttxx{investigates cross-channel correlations for modeling high-order feature statistics.} 
Given the context features $\mathcal{F}_{context}$ and object proposals (\ie, box and object queries), we first extract object features $\mathcal{F}_{obj}\in R^{ H\times W\times C}$ using ROIAlign~\cite{he2017mask}, where $H$, $W$, and $C$ indicate the height, width, and channel dimension of the features.
We then split $\mathcal{F}_{obj}$ along the channel dimension to generate channel-wise features \ttxx{$\mathcal{F}_{obj} = [\mathcal{F}_{obj}^{1}, \mathcal{F}_{obj}^{2}, ..., \mathcal{F}_{obj}^{K}]$}
$\in R^{C\times K}$, where $K = W\times H$ and $k \in [1, K]$, for computing the covariance normalization matrix $\mathcal{M}$ as:
\begin{equation}
    \centering
    \mathcal{M} = \left (\frac{1}{K}\sum_{k=1}^{K} \Big (\mathcal{F}_{obj}^{k}-\mu \Big )\Big (\mathcal{F}_{obj}^{k}-\mu \Big )^T \right )^\alpha \in R^{C\times C},
\end{equation}
where $\mu = \left (\frac{1}{K}\sum_{k=1}^{K}\mathcal{F}_{obj}^{k} \right ) \in R^{1 \times K}$ is the channel-wise mean vector. $\alpha$ is a normalization hyper parameter.
We then compute the global covariance pooling $\mathcal{GCP}$ on $\mathcal{M}$ as:
\begin{equation}
    \centering
    s_c = \mathcal{GCP}(\mathcal{M}_c) = \frac{1}{C}\sum_{c=1}^{C}\mathcal{M}_c ,
\end{equation}
where $\mathcal{S}=[s_1, s_2, ..., s_C] \in R^{C \times 1}$ is the high-order channel-wise feature statistics, which encodes objects and their relations to both local and global contexts.
%

\vspace{1mm}
\noindent{\bf Dynamic Rectifying Functions} 
\ttxx{further dynamically reallocate attentions to each feature channel  based on the channel-wise statistics $\mathcal{S}$.} It first learns to reweight each $s_c$ as:
%
%
\begin{equation}
\begin{split}
    \hat{s}_{c} =\max\{a_{c}^1(\mathcal{S})s_c+b_{c}^{1}(\mathcal{S}), a_{c}^2(\mathcal{S})s_c+b_{c}^{2}(\mathcal{S})\}, \label{eq:3}
\end{split}
\end{equation}  
where $a_{c}^1$, $b_{c}^1$, $a_{c}^2$, and $b_{c}^2$ are learnable parameters, which form two sets of coefficients (\ie, ($a_{c}^1$, $b_{c}^1$) and ($a_{c}^2$, $b_{c}^2$)) to formulate two piece-wise functions for updating $s_c$ to $\hat{s}_{c}$.
Since the max operation in Eq.~\ref{eq:3} is not differentiable, it is reformulated as:
\begin{gather*}
        [\Delta a_{1:C}^1,\Delta a_{1:C}^2,\Delta b_{1:C}^1,\Delta b_{1:C}^2]  \in R^{4\times C} = \notag \\
        2\sigma(FC_{R/C \to 4\times C}(ReLU(FC_{C \to R/C}(\mathcal{S})))) - 1, \tag{4} \label{eq:4} \\
        and, a_{1:C}^1 = 1 + \lambda_a \Delta a_{1:C}^1,\  a_{1:C}^2 =  \lambda_a \Delta a_{1:C}^2, \\
        b_{1:C}^1 =  \lambda_b \Delta b_{1:C}^1,\    b_{1:C}^2 = \lambda_b \Delta b_{1:C}^2, \tag{5}  \label{eq:5}
\end{gather*}
where $FC_{C \to R/C}$ is a fully connected layer, which changes the feature dimension from $C$ to $R/C$. $\sigma$ is a Sigmoid function. $\lambda_{a}$, $\lambda_{b}$ are hyper-parameters that are set to 1 and 0.5. $a_{1:C}^1 = [a_1^1, a_2^1, ..., a_C^1] \in R^{1\times C}$. $a_{1:C}^2$, $b_{1:C}^1$, and $b_{1:C}^2$ are defined similar to $a_{1:C}^1$.

Note that both $a$ and $b$ are related not only to the input $s_c$, but also to others $s_{i\neq c} \in \mathcal{S}$ via two $FC$ layers.
Hence, we establish the cross-channel correlations and compute the enhanced $\hat{\mathcal{F}}_{obj}$ as:
\begin{equation}
    \hat{\mathcal{F}}_{obj} = \mathcal{F}_{obj}\diamond \hat{S}, \tag{6}
\end{equation}
where $\hat{S}=[\hat{s_1},\hat{s_2},...,\hat{s_c}] \in R^{C\times 1}$, and $\diamond$ denotes channel-wise multiplication.

Given the globally pooled high-order channel-wise feature statistics, the dynamic rectifying function refines it via linear functions described in Eq.~\ref{eq:3}, where slopes and intercepts (\eg, ($a_{c}^1$, $b_{c}^1$)) of the linear functions are adaptively learned via Eq.~\ref{eq:4} and~\ref{eq:5}.
\ttxx{We leverage two sets of slopes and intercepts to help adjust negative/positive channel responses separately.}
In this way, fine-grained attentions can be adjusted via the slopes and intercepts before they are assigned to the feature channels, to accommodate for the saliency ranking task.
%


\subsection{Object-Context-Object Relation Module}

The OCOR module aims to model the spatial attention of the human visual system for learning how to prioritize regions.
%
%
%
To this end, we first encode the object-context relation based on the enhanced object representation of the SOS module, and then establish the bi-directional object-context-object relation, to mimic the spatial scanning process of humans.

\vspace{1mm}
\noindent{\bf Formulation of object-context relation.}
\ttxx{After the SOS module, we obtain $N$ object features $\hat{\mathcal{F}}_{obj(n)} \in R^{H \times W \times C}$.}
%
We then rescale $\mathcal{F}_{context}$ to have the same spatial size ($H \times W$) as $\hat{\mathcal{F}}_{obj(n)}$, and establish the object-context relation by concatenating each $\hat{\mathcal{F}}_{obj(n)}$ with $\mathcal{F}_{context}$ to form a series of object-context relation features $\{\mathcal{F}_{o \leftrightarrow c(n)}\}_{i=1}^{N} \in R^{H \times W \times 2C}$ for the $N$ objects.

\begin{figure}
    \centering
    \includegraphics[width=.45\textwidth]{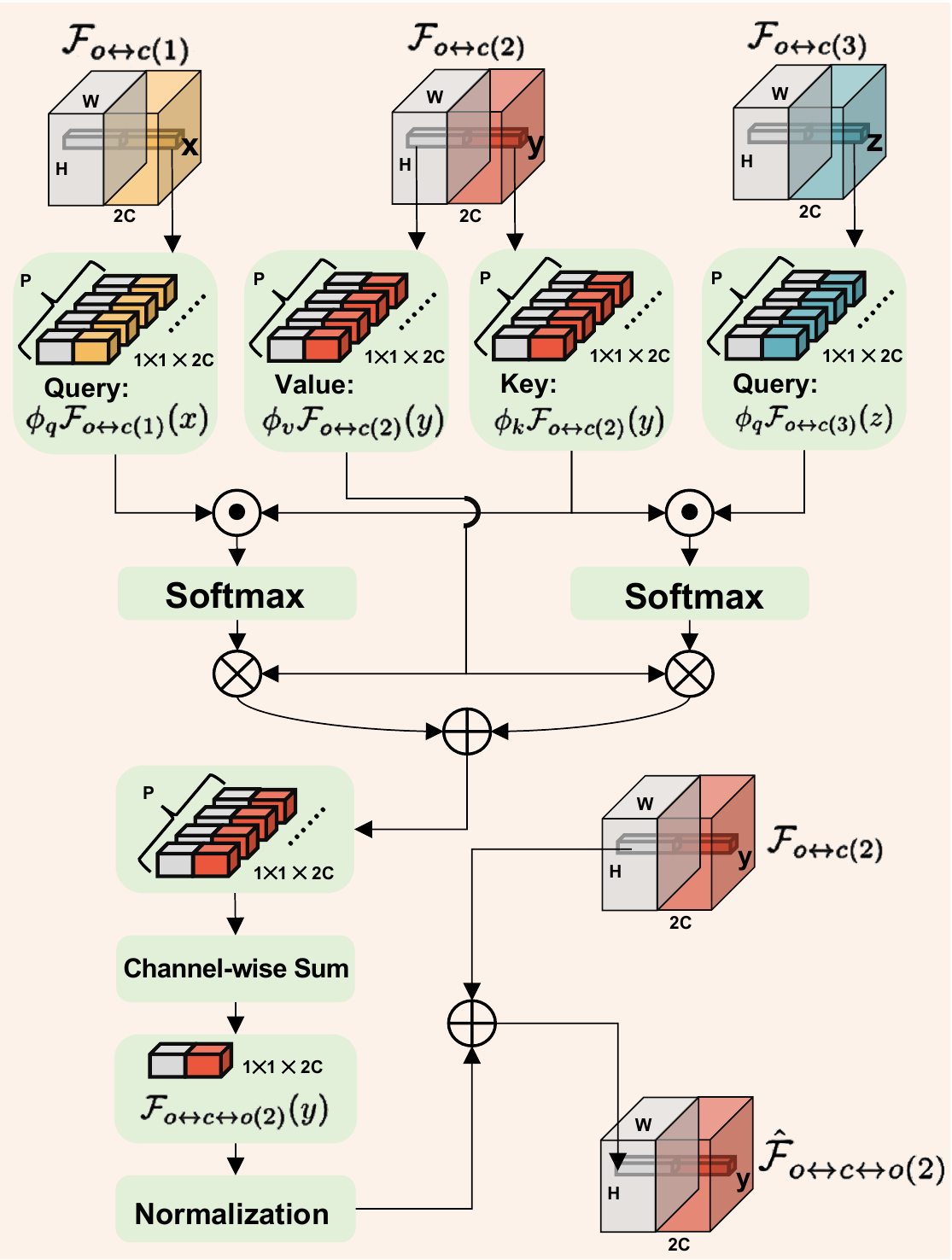}
    \caption{\ttxx{Structure of our OCOR module.}}
    \label{fig:ocor}
    \vspace{-6mm}
\end{figure}

\vspace{1mm}
\noindent{\bf Formulation of object-context-object relation.} Based on the object-context relations, we build bi-directional object-context-object relations $\{\mathcal{F}_{o\leftrightarrow c\leftrightarrow o(n}\}_{n=1}^{N}$ to model how attentions shift from region to region, accompanied with objects interacting with objects through contexts.
Specifically, we exploit groups of linear projections to compute long-range interactions between different object-context relation features, inspired by the multi-head attention mechanism~\cite{vaswani2017attention}.
We denote $i,j \in N$ as two different objects, $x, y \in K$ as spatial locations, and $\phi$ as a linear projection function. We utilize $P$ distinct projection functions, $\phi^{p},\ p \in [1,P]$, to obtain Key $k$, Query $q$, and Value $v$ embedding of the object-context relation features.
For example, $\phi_v^p{\mathcal{F}_{o\leftrightarrow c(i)} (x)}$ denotes the value embedding $v$ of object-context relation of object $i$ at location $x$ using the $p$-th linear projection function.
The interaction can be modeled as:
\begin{equation}
Att_{\langle i,j\rangle}^{p}(x,y) = \varrho (A_{\langle i,j\rangle}^{p}(x,y)) \times \phi_v^{p}(\mathcal{F}_{o\leftrightarrow c(i)} (x)), \tag{7} \label{eq:7}
\end{equation}
where $Att_{\langle i,j\rangle}^{p}(x,y)$ measures how object $i$ with its spatial context $x$ interacts with object $j$ with its spatial context $y$. $\varrho$ is the weighting function (\ie, Softmax) to compute the importance of the embedding of the object-context representation, and $A_{\langle i,j\rangle}^{p}(x,y)$ encodes the context information of $\mathcal{F}_{o\leftrightarrow c(i)}(x)$. They are calculated as:
\begin{equation}
    \varrho (A_{\langle i,j\rangle}^{p}(x,y)) = \frac{e^{A_{\langle i,j\rangle}^{p}(x,y)}}{\sum_{p^{'} \in P}e^{A_{\langle i,j\rangle}^{p^{'}}(x,y)}}, \tag{8} \label{eq:8}
\end{equation}
\begin{equation}
    A_{\langle i,j\rangle}^{p}(x,y) = \phi_k^{p}(\mathcal{F}_{o\leftrightarrow c(i)} (x))^{T} \cdot \phi_q^{p}(\mathcal{F}_{o\leftrightarrow c(j)} (y)),\tag{9} \label{eq:9}
\end{equation}
where $\cdot$ is the dot production between two feature vectors.
We compute the object-context-object relation features as:
\begin{equation}
    \mathcal{F}_{o\leftrightarrow c \leftrightarrow o (i)}(x) = \sum_{j=1,j\neq i}^{N}\sum_{y=1}^{K}\sum_{p=1}^{P} Att_{\langle i,j\rangle}^{p}(x,y). \tag{10}
\end{equation}
Finally, we normalize $\mathcal{F}_{o\leftrightarrow c \leftrightarrow o}$ and fuse it with the input $\mathcal{F}_{o\leftrightarrow c}$ to produce the final output $\mathcal{\hat{F}}_{o\leftrightarrow c \leftrightarrow o}$.
Figure~\ref{fig:ocor} shows the structure of our OCOR module.

\subsection{Learning of Saliency Rank}
To learn saliency ranking based on our SOS and OCOR modules,
we formulate the ranking step as a multi-stage query-based detection process, and follow~\cite{fang2021queryinst} to initialize with $T$ query stages ($1\leq t \leq T$). The advantage of this multi-stage strategy is that the box and object queries ($q_{box}^{t}$, and $q_{obj}^{t}$), and the corresponding object features $\mathcal{F}_{obj}^{t}$ can be improved stage by stage. At each stage, we perform the following three sub-tasks.
%
%
First, we extract the object features of the current stage $t$ (\ie, $\mathcal{F}_{obj}^{t}$) from the box query of stage $t-1$ (\ie, $q_{box}^{t-1}$) as:
\begin{equation}
\mathcal{F}_{obj}^{t} = \mbox{ROIAlign}(\mathcal{F}_{context}, q_{box}^{t-1}). \tag{11}
\end{equation}
Second, we obtain the improved object features $\mathcal{\Tilde{F}}_{obj}^{t}$ and object query $q_{obj}^{t}$ as:
\begin{equation}
\mathcal{\Tilde{F}}_{obj}^{t},\ q_{obj}^t = H^t_{rank} \left(\begin{matrix} \underbrace{ f_{OCOR}(f_{SOS}(\mathcal{F}_{obj}^{t})) } \\ \hat{\mathcal{F}}_{o \leftrightarrow c \leftrightarrow o}^{t} \end{matrix},\ f_{SA}(q_{obj}^{t-1}) \right), \tag{12}
\end{equation}
where $f_{SOS}$ and $f_{OCOR}$ are our SOS and OCOR modules, $f_{SA}$ is the multi-head self-attention \cite{vaswani2017attention}, and \kk{$H^t_{rank}$ is the ranking head at stage $t$}. 
%
Third, we update the box query using the box prediction branch $\mathcal{B}$:
\begin{equation}
q_{box}^{t} = \mathcal{B}_t(\mathcal{\Tilde{F}}_{obj}^{t}). \tag{13}
\end{equation}
\kk{At the final stage, the object features $\mathcal{\Tilde{F}}_{obj}^{T}$ are fed into the ranking head for predicting the final saliency ranks, and the mask head for predicting the salient object masks.}

\section{Experiments}

\subsection{Experimental Setups}
\noindent {\bf Datasets and Metrics.} Our experiments are conducted on the publicly available ASSR~\cite{siris2020inferring} and IRSR~\cite{liu2021instance} datasets. 
ASSR~\cite{siris2020inferring} ranks 5 salient objects per image based on the sequential eye gaze information. It provides 7,464, 1,436, and 2,418 images for training, validation, and testing, respectively.
IRSR~\cite{liu2021instance} considers both eye gaze sequences and the duration of eye gaze for labeling the saliency ranks. It also manually filters out inappropriate annotations. In IRSR, each image involves at most 8 salient objects with ranks. It contains 8,988 images, which are divided into 6,059 images for training and 2,929 images for test.

We adopt three metrics, \ie, Salient Object Ranking (SOR) \cite{islam2018revisiting, siris2020inferring}, Segmentation-Aware SOR (SA-SOR) \cite{liu2021instance}, and Mean Absolute Error (MAE), to evaluate our method. 
SOR computes the Spaearman's rank-order correlation between the prediction and ground truth of the saliency ranking order. This metric indicates the prediction quality of relative saliency among objects rather than a particular saliency ranking number for an object.
SA-SOR calculates the Pearson correlation between the prediction and ground truth of saliency ranks. It also penalizes the errors of detecting non-salient objects and false ranking.
%
%
MAE measures averaged per-pixel differences between the prediction and ground truth of saliency map.
\tx{Hence, it also helps measure the ranking quality of an object depending on its overlapping degree with the ground truth.}

\vspace{1mm}
\noindent {\bf Implementation Details.} Our network is built upon the multi-stage query-based detector~\cite{fang2021queryinst, sun2021sparse}. Following their settings, the stage number $T$ is set to 6, and the query number $N$ is set to 100. We utilize the Swin transformer~\cite{liu2021swin} pretrained on ImageNet~\cite{krizhevsky2012imagenet} as our backbone. \ttxx{$H$ and $W$ are initially set to 7 in the SOS module, and then reduced to 2 in the OCOR module. We set the $\alpha$ to $\frac{1}{2}$ and $C$ to $256$.} We experimentally set $R=4$ and $P=8$ for our SOS and OCOR modules, respectively.

%
We train our model for 60 epochs on each dataset. The learning rate starts from $2.5\times 10^{-5}$, and is divided by 10 at epochs 25 and 45. 
To optimize the model, we adopt AdamW optimizer with a $1\times 10^{-4}$ weight decay.
Images are resized to $800\times 800$ resolution. We use random flip for data augmentation.
The batch size is set to 16 for training on 4 Tesla GPU cards.
We train the ranking head with set prediction loss \cite{fang2021queryinst, sun2021sparse} and saliency ranking loss \cite{liu2021instance}, and the mask head with dice loss \cite{milletari2016v}. We test our model on a single Tesla GPU card with images of $800 \times 800$ resolution.


%


\subsection{Baselines}
\label{sec:43}
Since saliency ranking is a relatively new task with only four methods proposed,
to evaluate our method comprehensively, we design baseline methods for comparisons.
We find the Semantic Instance Segmentation task related to our task in that they have a similar output formulation (\ie,  object instance maps + corresponding semantic categories) to  our task (\ie, salient instance maps + saliency ranks).
Hence, we include eight representative semantic instance segmentation methods as baselines for comparisons.
\begin{itemize}
    \item Mask R-CNN \cite{he2017mask} is a  popular two-stage detector. All existing salient object ranking methods \cite{siris2020inferring, liu2021instance, fang2021salient} are built on it. Here, we have two versions: a ResNet-based Mask R-CNN, and a Swin-based Mask R-CNN.
    \item BlendMask \cite{chen2020blendmask} is a single-stage anchor-free method that integrates high-level task semantics (from top-down) and low-level finer details (from bottom-up). It provides a simpler baseline compared to the anchor-based detectors~\cite{he2017mask}.
    \item CenterMask \cite{lee2020centermask} is also a single-stage anchor-free network consisting of a channel attention-based backbone (VoVNet) and a spatial attention-based mask head for extracting informative semantics and suppressing background noise, respectively.
    \item SOLO \cite{wang2020solo} proposes to classify and detect an instance at each location of the feature grid. They directly combine coordinate information with candidate object features so that the detection can be more sensitive to the object centroids and boundaries.
    \item Cascade R-CNN \cite{cai2018cascade} and QueryInst \cite{fang2021queryinst} detect instances in a progressive manner. They utilize cascaded detectors in a network to refine the detection results stage-by-stage. 
    \item CBNetV2 \cite{liang2021cbnetv2} proposes to connect parallel backbones in a top-down pathway. The higher-level features, with coarser space information and finer semantic information, of the former backbone can be used to enrich lower-level features of the later backbone. 
\end{itemize}

To adapt these methods for saliency ranking, we modify their output layers according to the amounts of saliency ranks in the saliency ranking datasets. We train these methods using the saliency ranking loss \cite{liu2021instance}.



\newcolumntype{\thc}{!{\vrule width 1pt}}

\begin{table*}[!t]\small
		\centering
		\caption{Quantitative comparison with 4 state-of-the-art saliency ranking methods, 4 salient object detection methods used by~\cite{siris2020inferring} for comparison, and
		our 8 baselines in Section~\ref{sec:43}. We show their original tasks and backbones used (\ie, ResNet~\cite{he2016deep}, VoVNet~\cite{lee2020centermask}, and Swin~\cite{liu2021swin}) in the 2$nd$ and 3$rd$ columns. SID, SOD, SIS, and SR represent salient instance detection, salient object detection, semantic instance segmentation, and saliency ranking, respectively. Methods with $\dag$ denotes that their  results are copied from their original papers. - denotes missing results due to the lacking of publicly available implementations/results. Best performances are marked in {\bf bold}.}
		\begin{tabular}{c c c c c c c c c}\toprule[1.5pt]
			\multirow{3}*{\bf Method}&\multirow{3}*{\bf Original Task}&\multirow{3}*{\bf Backbone} &\multicolumn{6}{c}{\bf Benchmark Datasets and Evaluation Metrics}\\\cline{4-9}
			&&&\multicolumn{3}{c}{\textbf{ASSR Test Set}~\cite{siris2020inferring}}&\multicolumn{3}{c}{\textbf{IRSR Test Set}~\cite{liu2021instance}}\\\cline{4-9}
			&&&\small{\textbf{SA-SOR}$\uparrow$}&\small{\textbf{SOR}$\uparrow$}&\small{ \textbf{MAE}$\downarrow$}&\small{\textbf{SA-SOR}$\uparrow$}&\small{\textbf{SOR}$\uparrow$}&\small{\textbf{MAE}$\downarrow$}\\\midrule[1.0pt]
			S4Net$^\dag$ \cite{fan2019s4net}&SID&ResNet-50&-&0.891&0.150&-&-&-\\
			BASNet$^\dag$ \cite{qin2019basnet}&SOD&ResNet-34&-&0.707&0.115&-&-&-\\
			CPD-R$^\dag$ \cite{wu2019cascaded}&SOD&ResNet-50&-&0.766&0.100&-&-&-\\
			SCRN$^\dag$ \cite{wu2019stacked}&SOD&ResNet-50&-&0.756&0.116&-&-&-\\
			RSDNet~\cite{islam2018revisiting}&SR&ResNet-101&0.499 & 0.717 & 0.158 & 0.460 & 0.735 & 0.129\\
			ASSR~\cite{siris2020inferring}&SR&ResNet-101& 0.667 & 0.792 & 0.101 & 0.388 & 0.714 & 0.125\\
			IRSR~\cite{liu2021instance}&SR&ResNet-50& 0.709 & 0.811 & 0.105 & 0.565 & 0.806 & 0.085\\
			SOR$^\dag$ \cite{fang2021salient}&\makecell[c]{SR and SIS \\ (joint learning)}&VoVNet-39&-&0.841&0.081&-&-&-\\
			CenterMask \cite{lee2020centermask}&SIS&VoVNet-99& 0.672 & 0.813 & 0.099 & 0.509 & 0.771 & 0.113\\
			SOLO \cite{wang2020solo}&SIS&ResNet-101& 0.655 & 0.805 & 0.112 & 0.499 & 0.745 & 0.126\\
			BlendMask \cite{chen2020blendmask}&SIS&ResNet-101& 0.694 & 0.822 & 0.094 & 0.531 & 0.785 & 0.098\\
			Mask R-CNN \cite{he2017mask}&SIS&ResNet-101& 0.632 & 0.739 & 0.123 & 0.480 & 0.699 & 0.137\\
			Mask R-CNN \cite{he2017mask}&SIS&Swin-L& 0.643 & 0.750 & 0.118 & 0.489 & 0.712 & 0.128\\
			Cascade R-CNN \cite{cai2018cascade}&SIS&Swin-L& 0.699 & 0.816 & 0.100 & 0.520 & 0.766 & 0.105\\
			QueryInst \cite{fang2021queryinst}&SIS&Swin-L& 0.715 & 0.837 & 0.095 & 0.542 & 0.799 & 0.087\\
			CBNetV2 \cite{liang2021cbnetv2}&SIS&\small{Cascaded Swin-L}& 0.704 & 0.827 & 0.101 & 0.514 & 0.780 & 0.091\\
			Ours&SR&Swin-L& \textbf{0.738} & \textbf{0.904} & \textbf{0.078} & \textbf{0.578} & \textbf{0.834} & \textbf{0.079}\\\bottomrule[1.5pt]
		\end{tabular}
		\label{tab:main}
\end{table*}

\subsection{Main Results}
\label{sec:44}
We compare our method to 
\textbf{\underline{four}} existing saliency ranking methods: RSDNet \cite{islam2018revisiting}, ASSR \cite{siris2020inferring}, IRSR \cite{liu2021instance}, and SOR \cite{fang2021salient}; 
\textbf{\underline{four}} salient object detection methods used by~\cite{siris2020inferring} for comparison: S4Net \cite{fan2019s4net}, BASNet \cite{qin2019basnet}, CPD-R \cite{wu2019cascaded}, and SCRN \cite{wu2019stacked};
and our {\bf \underline{eight}} baselines from semantic instance segmentation: CenterMask \cite{lee2020centermask}, SOLO \cite{wang2020solo}, BlendMask \cite{chen2020blendmask}, ResNet-based Mask R-CNN \cite{he2017mask}, Swin-based Mask R-CNN \cite{he2017mask}, Cascade R-CNN \cite{cai2018cascade}, QueryInst \cite{fang2021queryinst}, and CBNetV2 \cite{liang2021cbnetv2}.

\noindent{\bf Quantitative Comparisons.}
Table~\ref{tab:main} shows the quantitative results. 
From the comparison on the ASSR testing set, we can see that S4Net performs the second best on SOR, but the second worst based on MAE. This is because missing objects are not penalized by SOR but are considered by MAE.
SOD-based methods perform relatively bad on SOR, as they do not have ability to rank object saliency.
Several SIS baselines perform well on the SR task.
%
Overall, the proposed method outperforms all methods compared by a large margin on all three metrics and on both testing sets.

\begin{figure*}[!t]
\centering
\begin{minipage}[t]{0.075\textwidth}
\centering
{{\scriptsize{Image}}}
\end{minipage}
\begin{minipage}[t]{0.075\textwidth}
\centering
{{\scriptsize{RSDNet\\ \vspace{-0.15cm} \cite{islam2018revisiting}}}}
\end{minipage}
\begin{minipage}[t]{0.075\textwidth}
\centering
{{\scriptsize{BlendMask\\ \vspace{-0.15cm} \cite{chen2020blendmask}}}}
\end{minipage}
\begin{minipage}[t]{0.075\textwidth}
\centering
{{\scriptsize{CenterMask\\ \vspace{-0.15cm} \cite{lee2020centermask}}}}
\end{minipage}
\begin{minipage}[t]{0.075\textwidth}
\centering
{{\scriptsize{SOLOv2\\ \vspace{-0.15cm} \cite{wang2020solo}}}}
\end{minipage}
\begin{minipage}[t]{0.075\textwidth}
\centering
{\scriptsize{Cascade\\ \vspace{-0.15cm} R-CNN \cite{wu2019cascaded}}}
\end{minipage}
\begin{minipage}[t]{0.075\textwidth}
\centering
{{\scriptsize{CBNetv2\\ \vspace{-0.15cm} \cite{liang2021cbnetv2}}}}
\end{minipage}
\begin{minipage}[t]{0.075\textwidth}
\centering
{{\scriptsize{QueryInst\\ \vspace{-0.15cm} \cite{fang2021queryinst}}}}
\end{minipage}
\begin{minipage}[t]{0.075\textwidth}
\centering
{{\scriptsize{ASSR \\ \vspace{-0.15cm} \cite{siris2020inferring}}}}
\end{minipage}
\begin{minipage}[t]{0.075\textwidth}
\centering
{{\scriptsize{IRSR\\ \vspace{-0.15cm} \cite{liu2021instance}}}}
\end{minipage}
\begin{minipage}[t]{0.075\textwidth}
\centering
{{\scriptsize{Ours}}}
\end{minipage}
\begin{minipage}[t]{0.075\textwidth}
\centering
{{\scriptsize{GT}}}
\end{minipage}
\\
\vspace{0.15cm}
\scriptsize{\underline{\textbf{(1) Examples of ranking salient objects based on their interactions about the activity context of a scene.}}}
\\
\vspace{0.08cm}
\scriptsize{\underline{Row 1: playing a video game, Row 2: flying a kite, Row 3: human-to-animal interaction, and Row 4: playing the baseball.}}
\\
\vspace{0.07cm}
\includegraphics[width=.075\textwidth]{quality_res/3/COCO_val2014_000000257370.jpg}
\centering
\includegraphics[width=.075\textwidth]{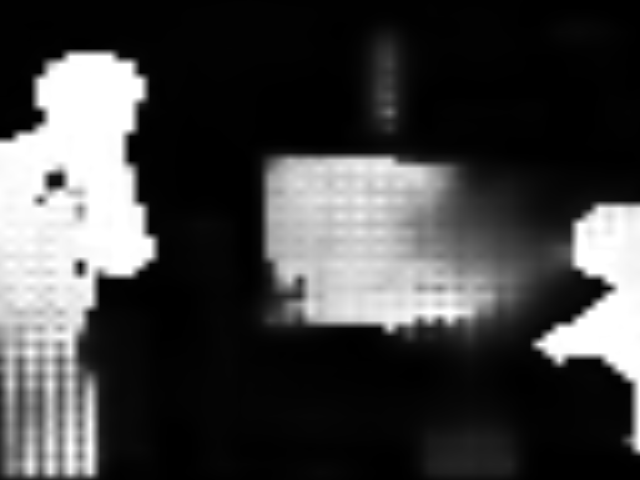}
\centering
\includegraphics[width=.075\textwidth]{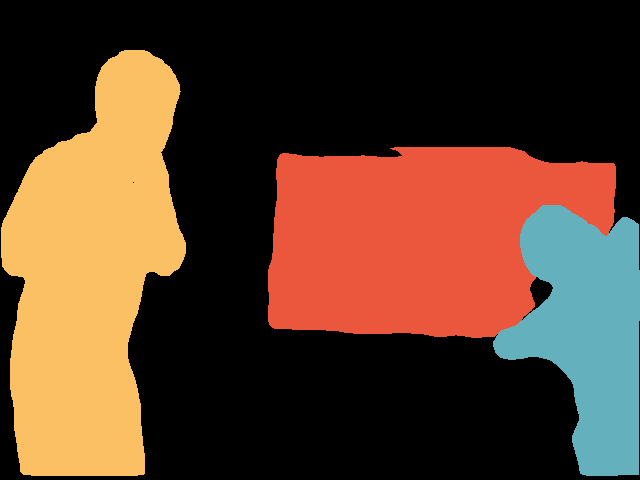}
\centering
\includegraphics[width=.075\textwidth]{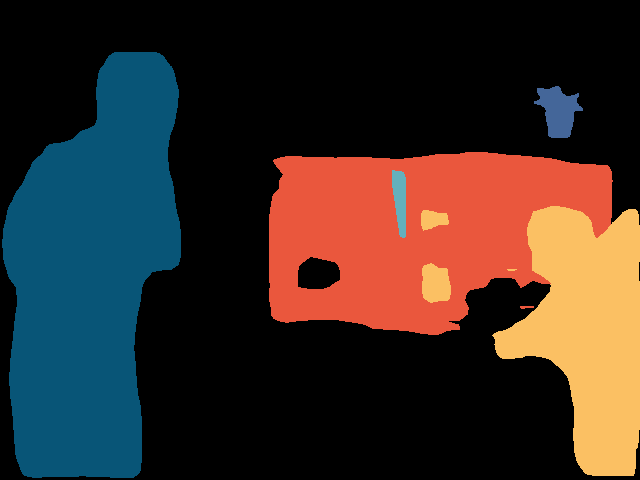}
\centering
\includegraphics[width=.075\textwidth]{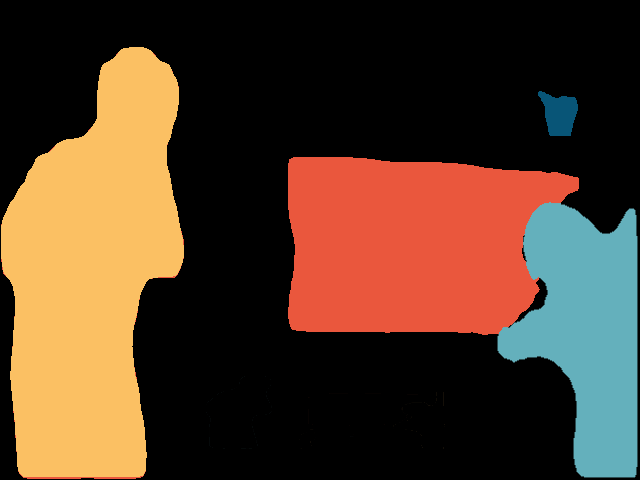}
\centering
\includegraphics[width=.075\textwidth]{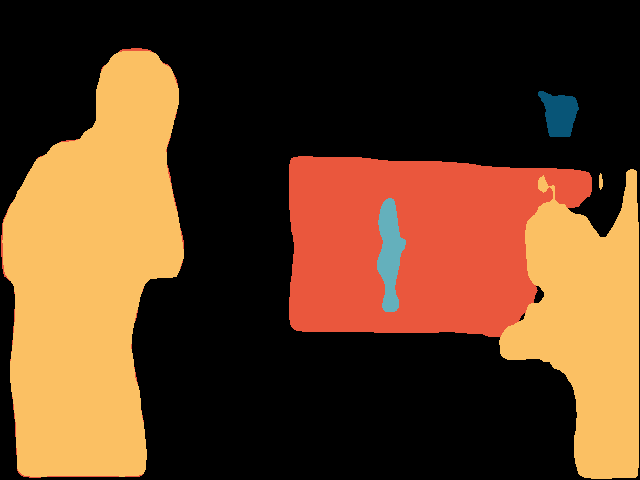}
\centering
\includegraphics[width=.075\textwidth]{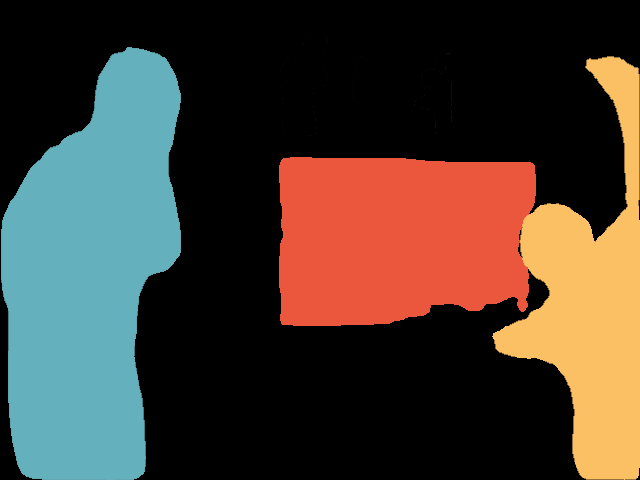}
\centering
\includegraphics[width=.075\textwidth]{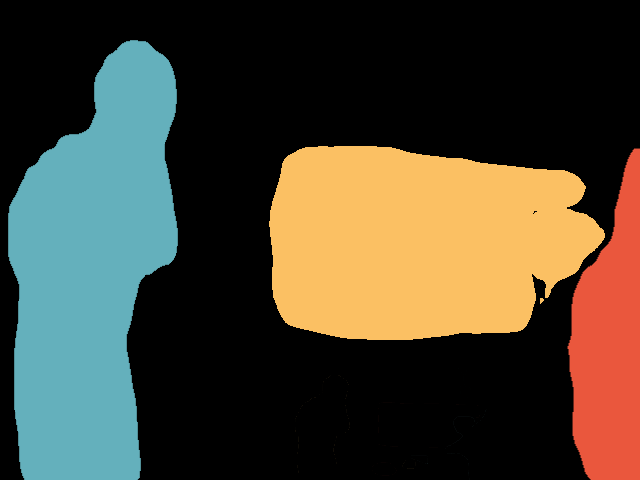}
\centering
\includegraphics[width=.075\textwidth]{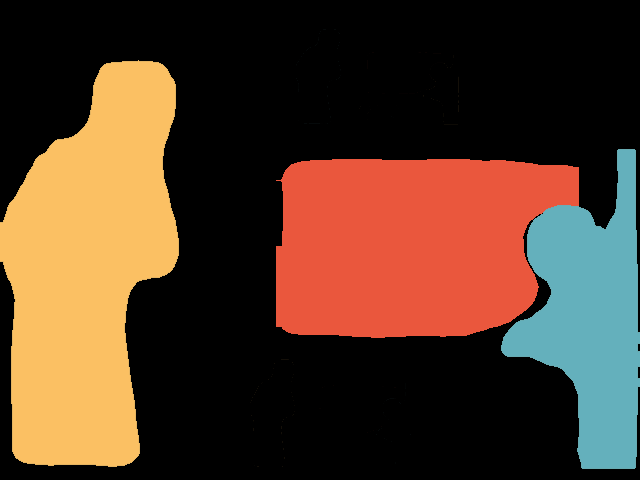}
\centering
\includegraphics[width=.075\textwidth]{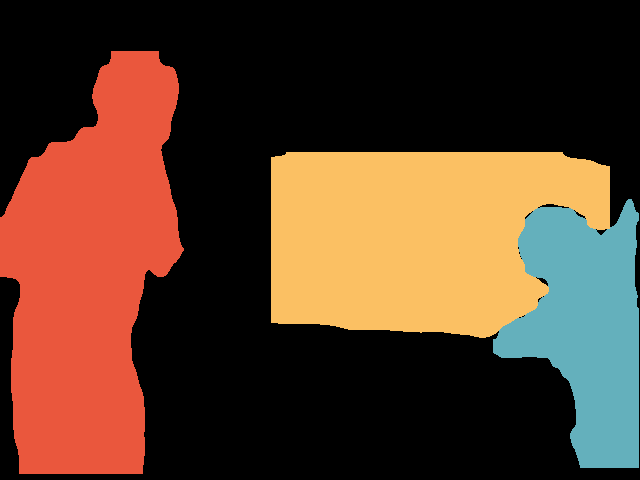}
\centering
\includegraphics[width=.075\textwidth]{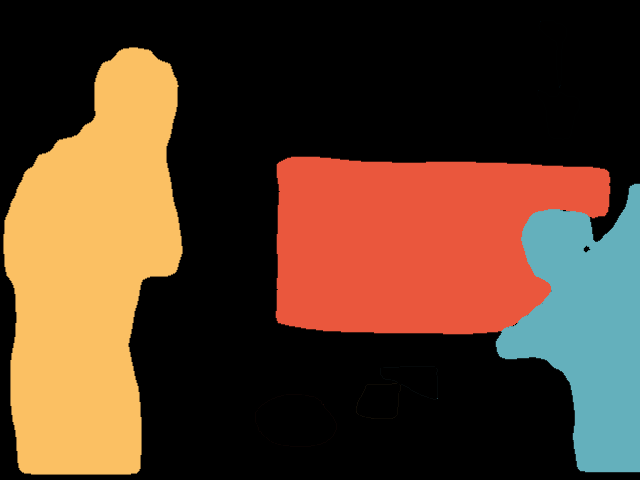}
\centering
\includegraphics[width=.075\textwidth]{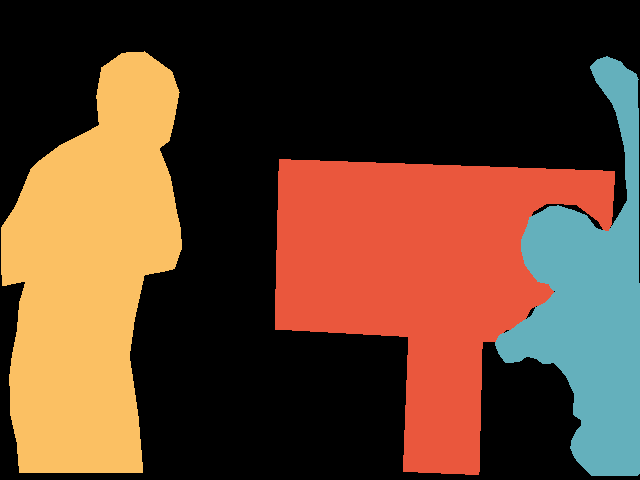}
\\
\vspace{0.07cm}
\includegraphics[width=.075\textwidth]{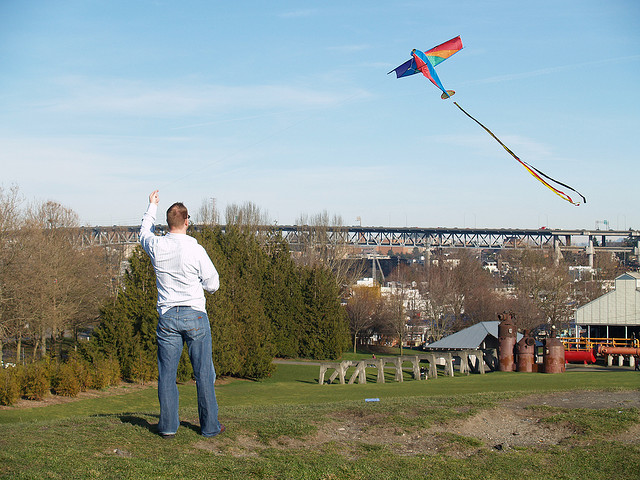}
\includegraphics[width=.075\textwidth]{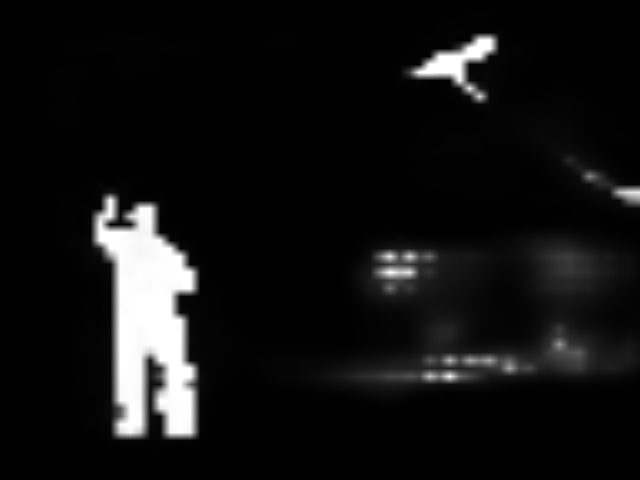}
\includegraphics[width=.075\textwidth]{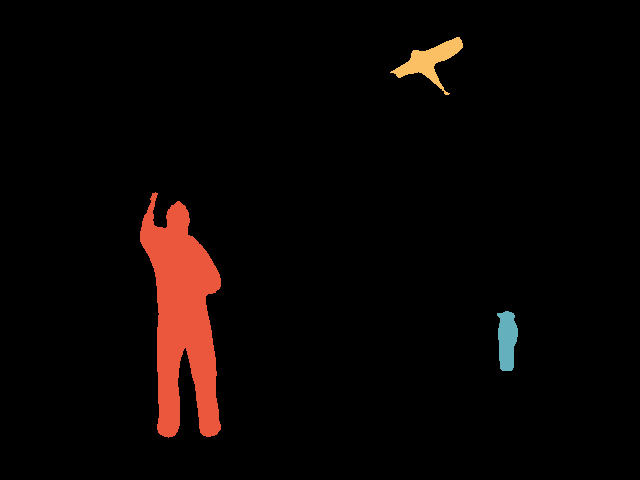}
\includegraphics[width=.075\textwidth]{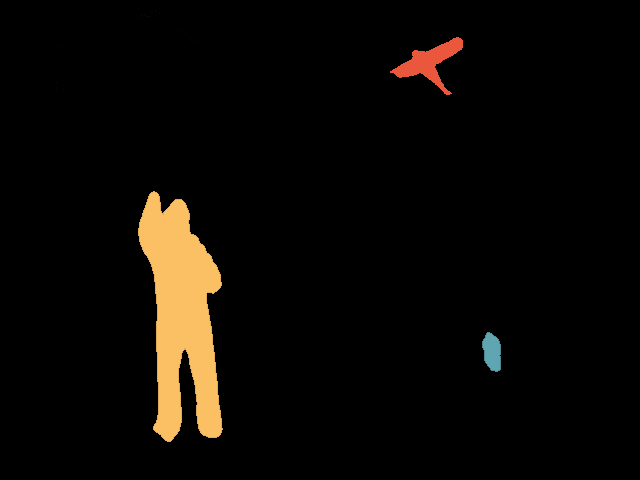}
\includegraphics[width=.075\textwidth]{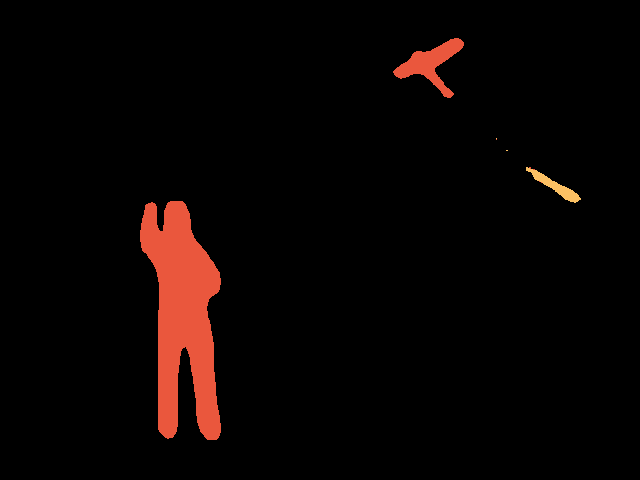}
\includegraphics[width=.075\textwidth]{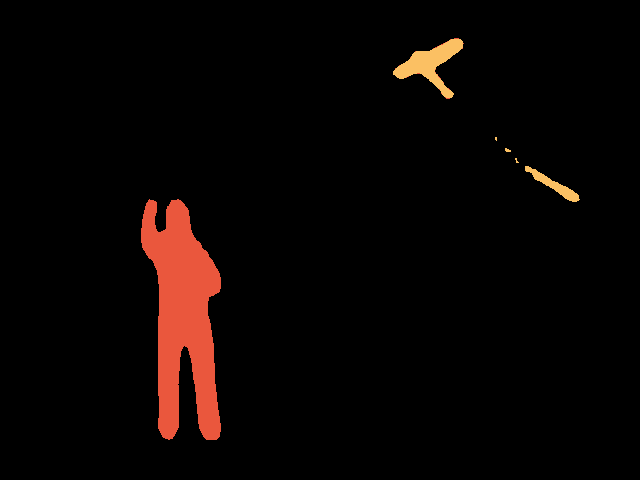}
\includegraphics[width=.075\textwidth]{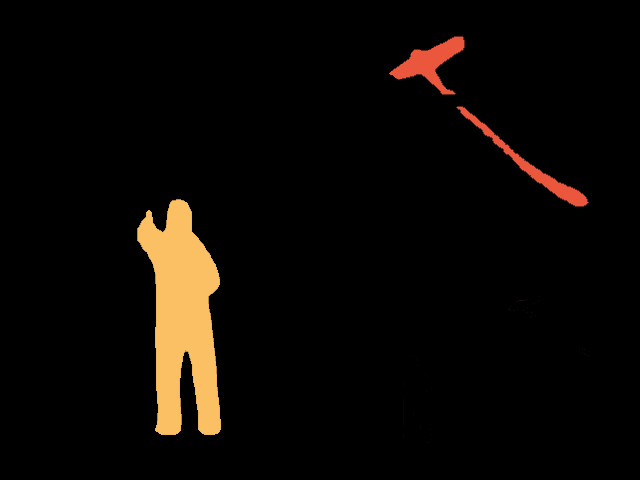}
\includegraphics[width=.075\textwidth]{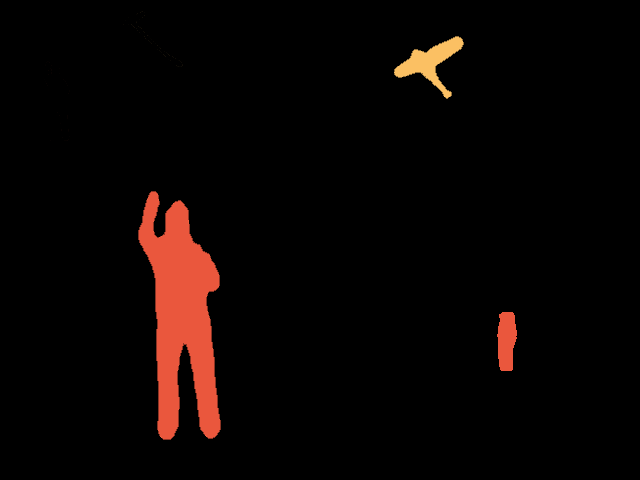}
\includegraphics[width=.075\textwidth]{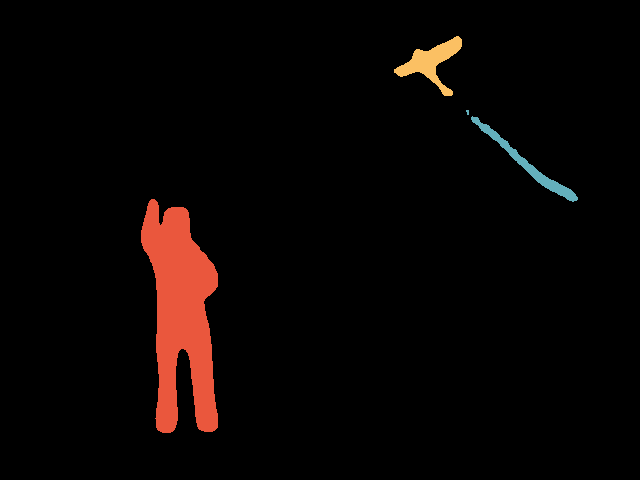}
\includegraphics[width=.075\textwidth]{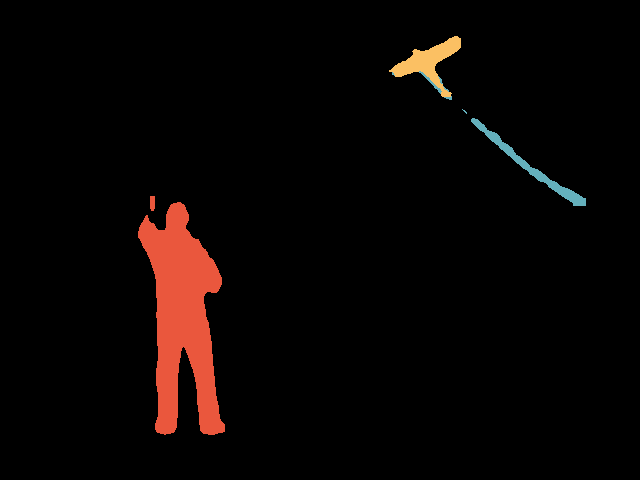}
\includegraphics[width=.075\textwidth]{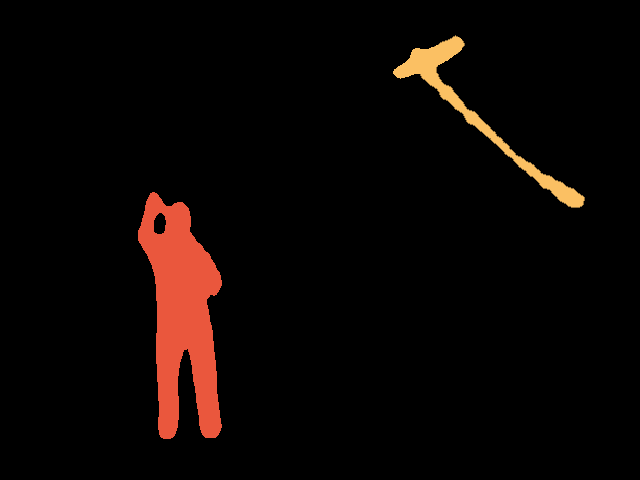}
\includegraphics[width=.075\textwidth]{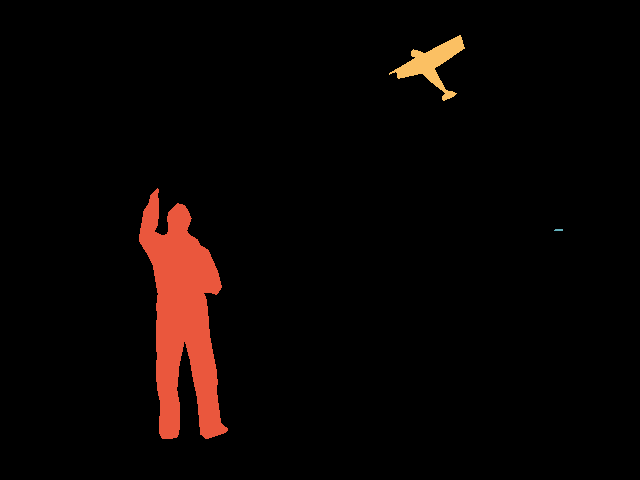}
\\
\vspace{0.07cm}
\includegraphics[width=.075\textwidth]{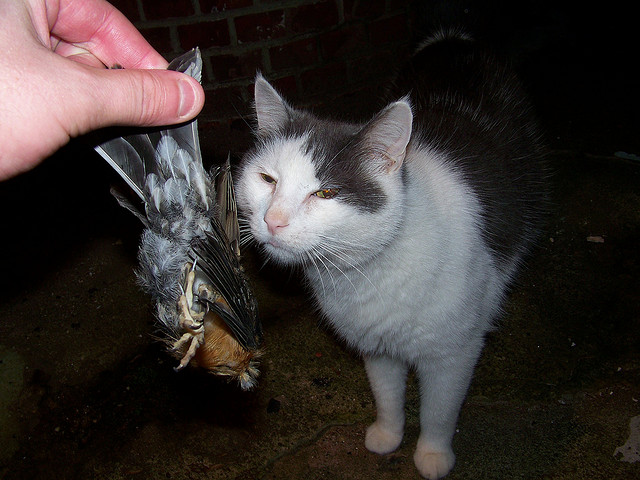}
\includegraphics[width=.075\textwidth]{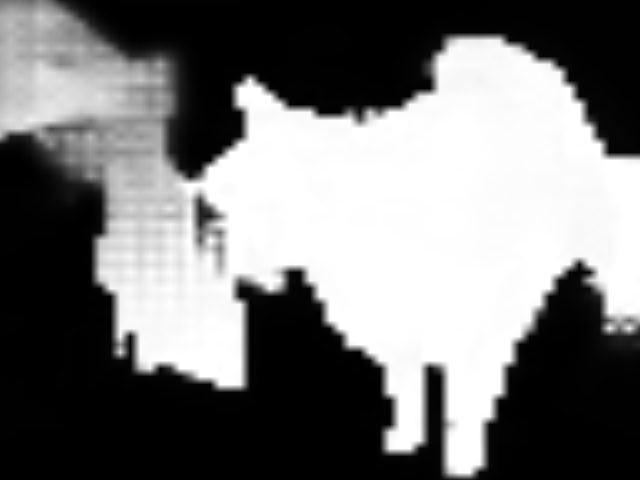}
\includegraphics[width=.075\textwidth]{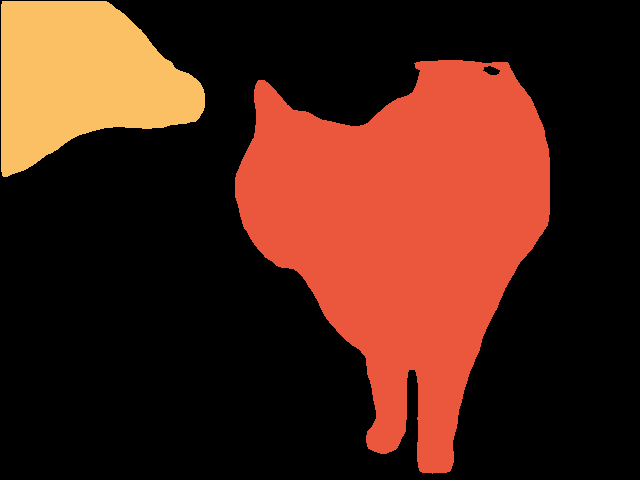}
\includegraphics[width=.075\textwidth]{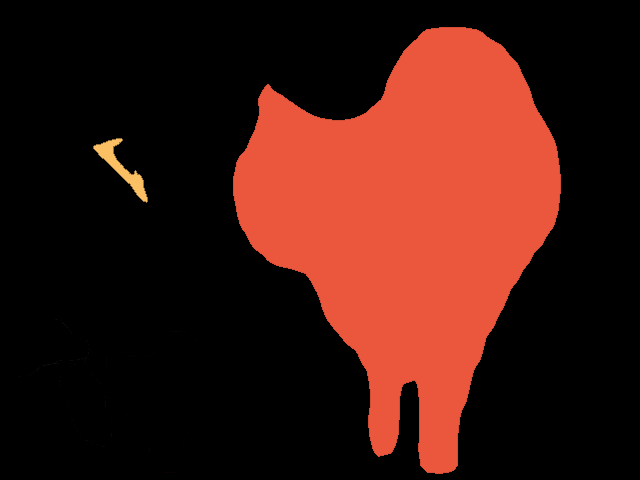}
\includegraphics[width=.075\textwidth]{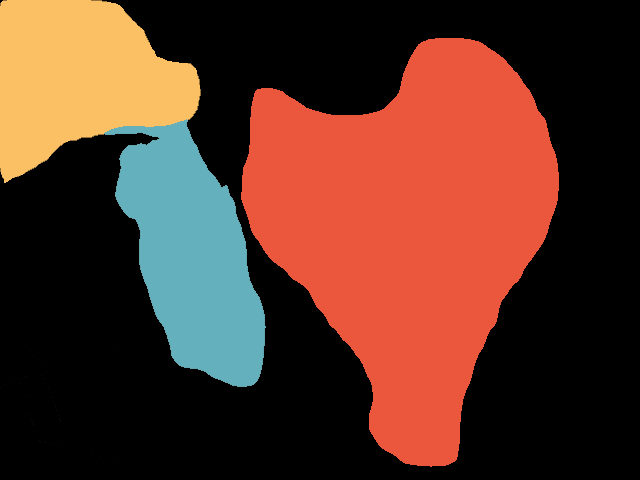}
\includegraphics[width=.075\textwidth]{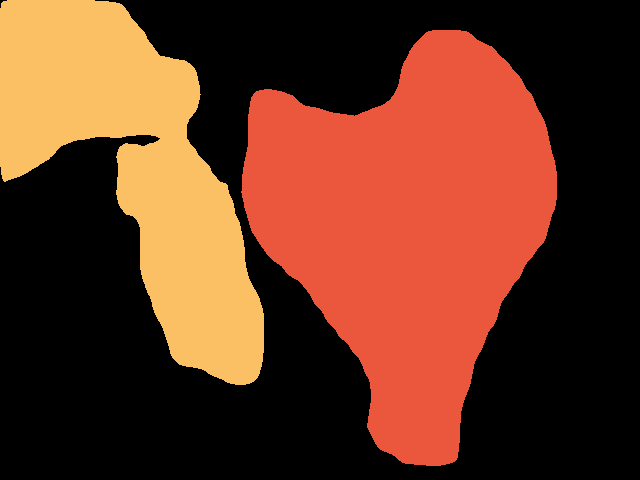}
\includegraphics[width=.075\textwidth]{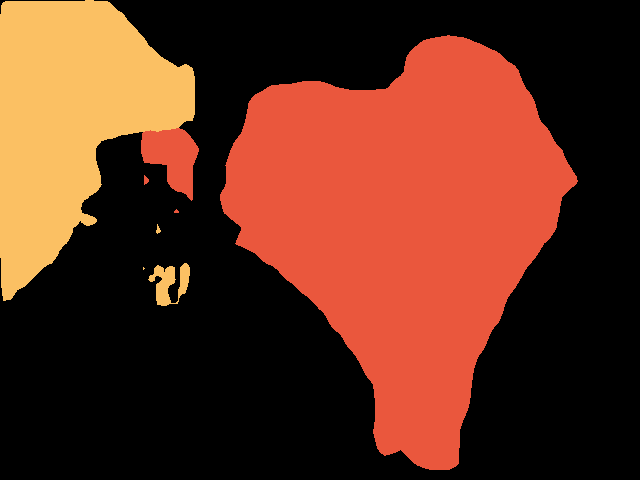}
\includegraphics[width=.075\textwidth]{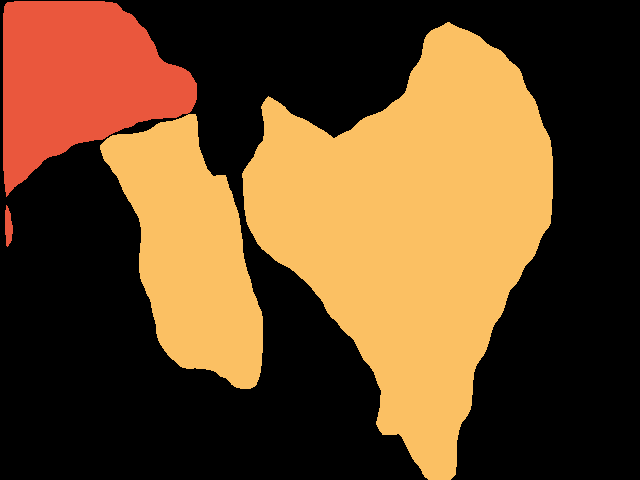}
\includegraphics[width=.075\textwidth]{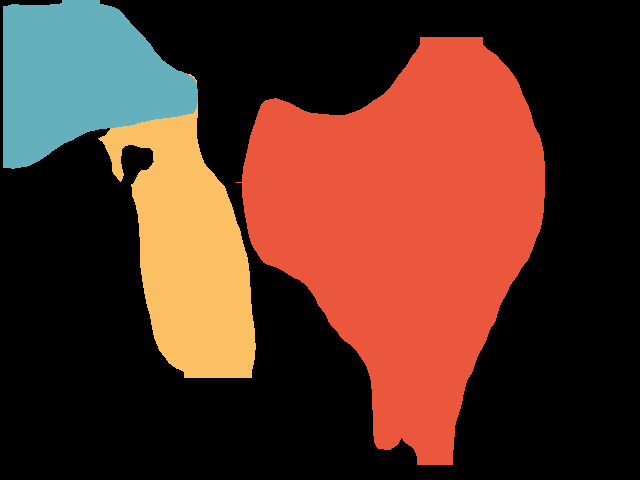}
\includegraphics[width=.075\textwidth]{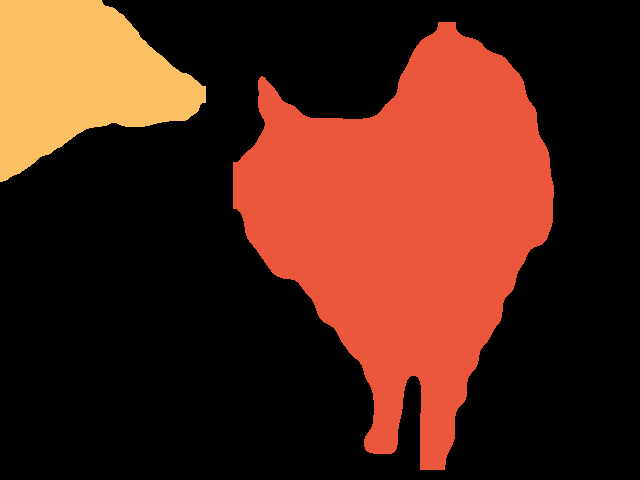}
\includegraphics[width=.075\textwidth]{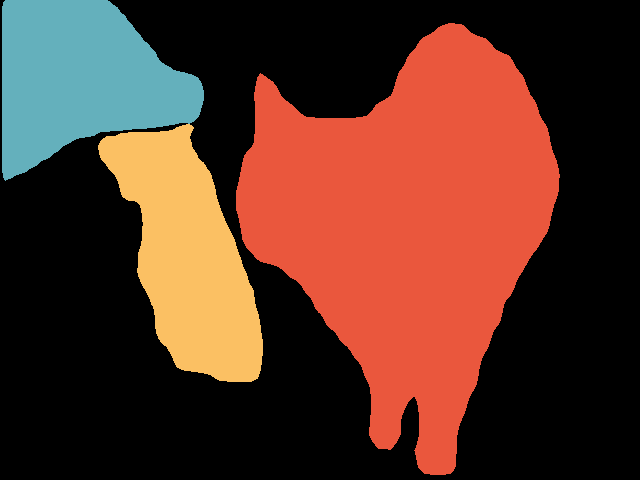}
\includegraphics[width=.075\textwidth]{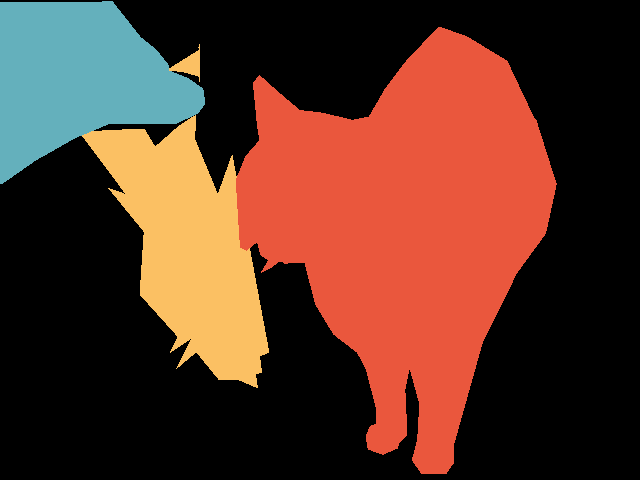}
\\
\vspace{0.07cm}
\includegraphics[width=.075\textwidth]{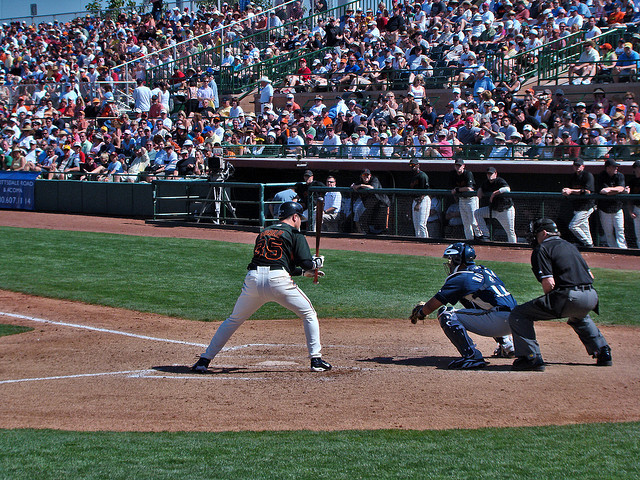}
\includegraphics[width=.075\textwidth]{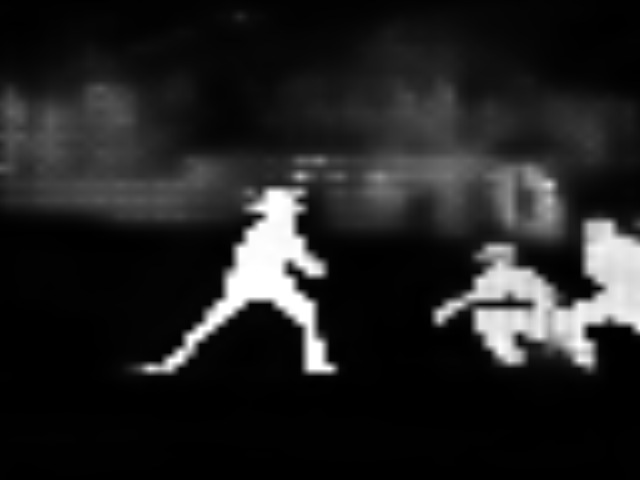}
\includegraphics[width=.075\textwidth]{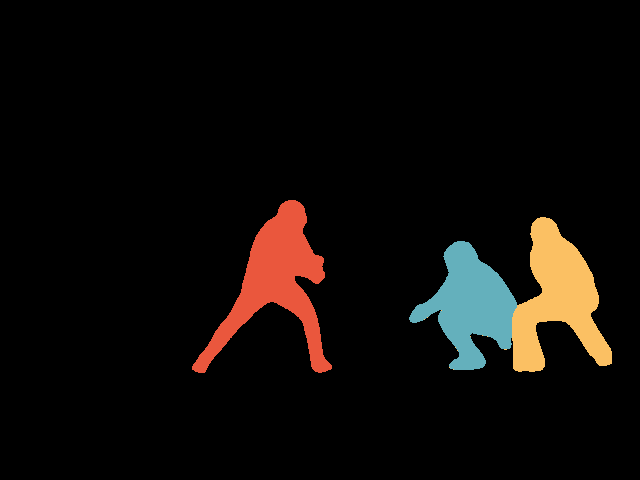}
\includegraphics[width=.075\textwidth]{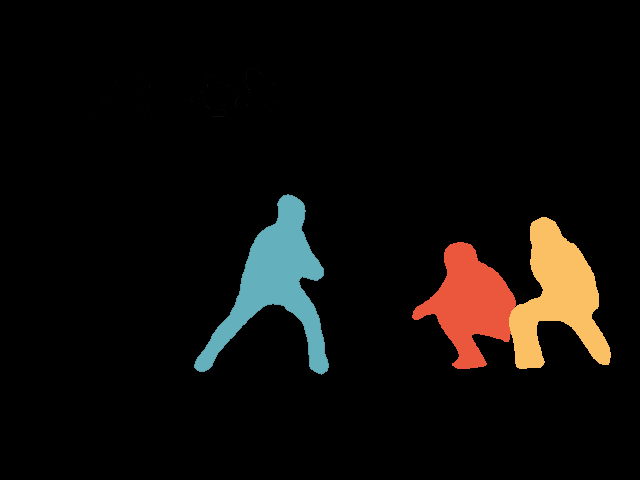}
\includegraphics[width=.075\textwidth]{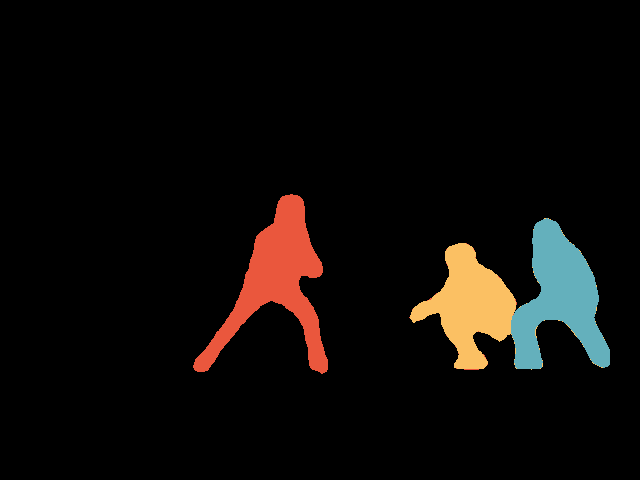}
\includegraphics[width=.075\textwidth]{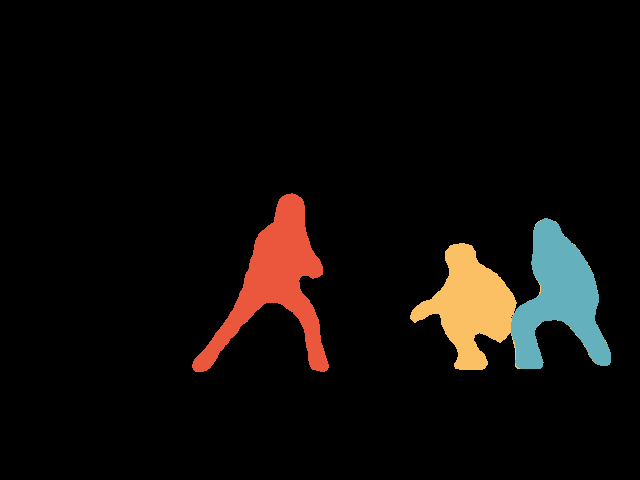}
\includegraphics[width=.075\textwidth]{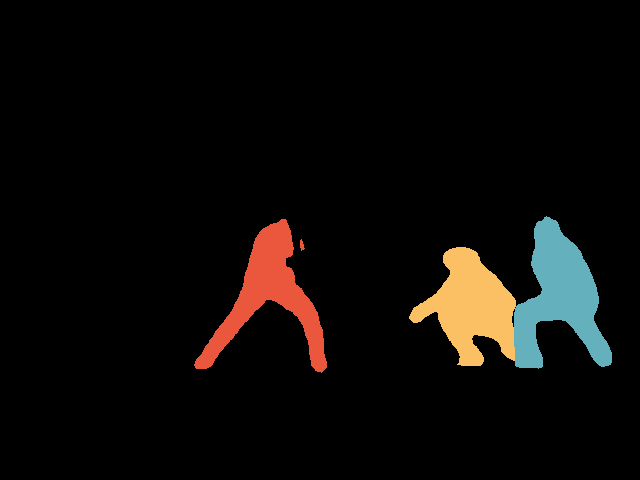}
\includegraphics[width=.075\textwidth]{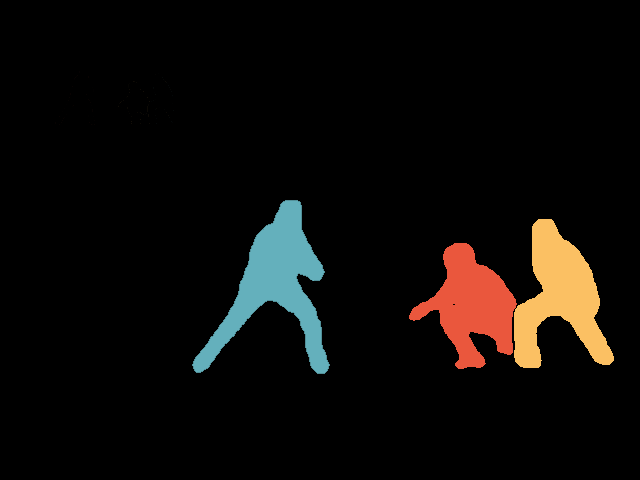}
\includegraphics[width=.075\textwidth]{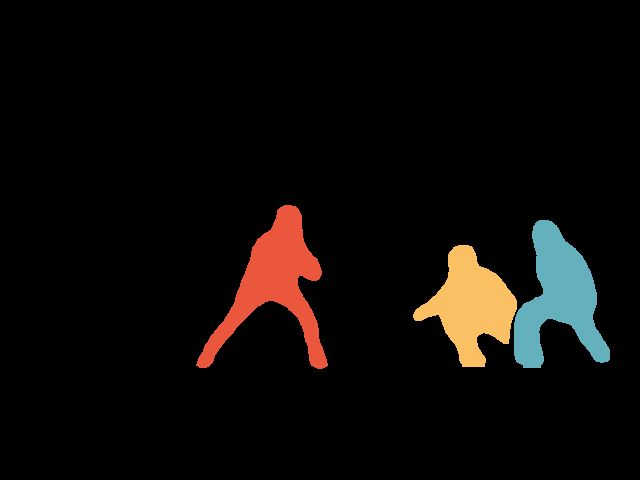}
\includegraphics[width=.075\textwidth]{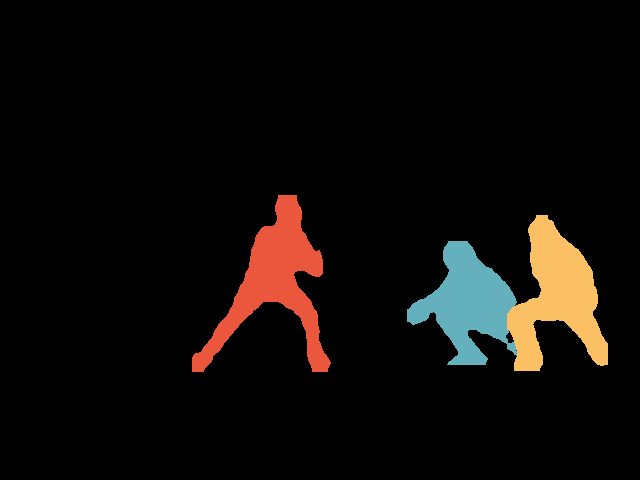}
\includegraphics[width=.075\textwidth]{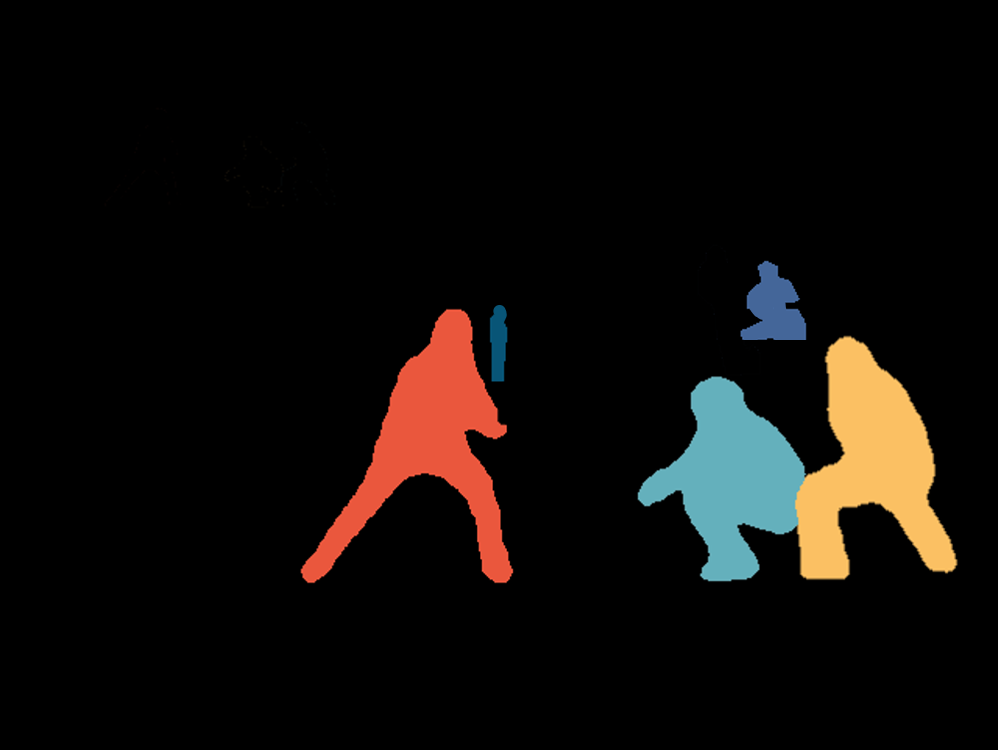}
\includegraphics[width=.075\textwidth]{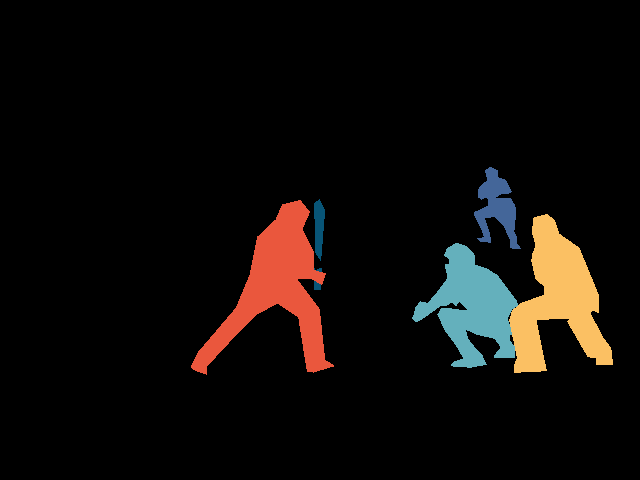}
\\
\scriptsize{\bf\underline{(2) Examples of ranking salient objects based on their correlated functionalities.}}
\\
\vspace{0.08cm}
\scriptsize{\underline{Row 5: computer in an office, and Row 6: food of a meal.}}
\\
\vspace{0.07cm}
\includegraphics[width=.075\textwidth]{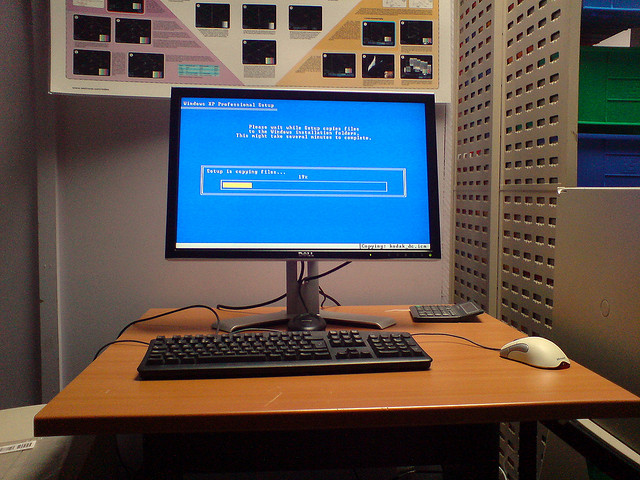}
\includegraphics[width=.075\textwidth]{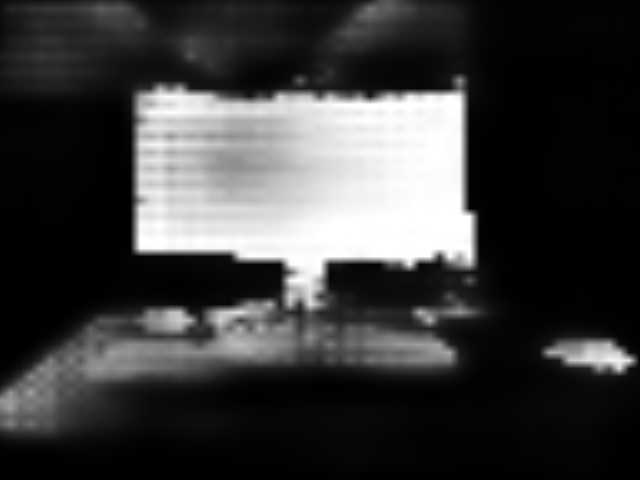}
\includegraphics[width=.075\textwidth]{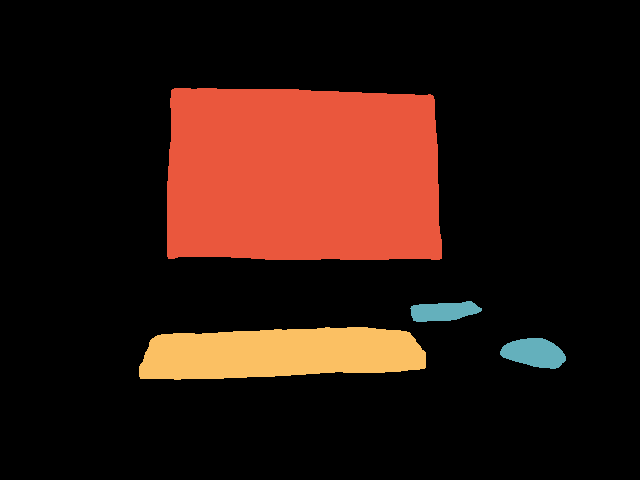}
\includegraphics[width=.075\textwidth]{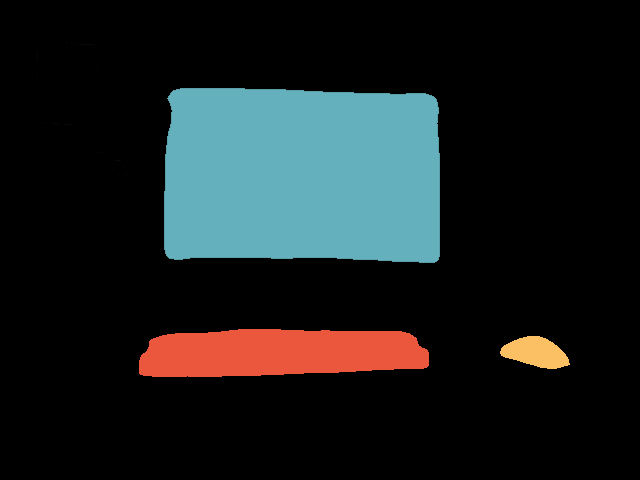}
\includegraphics[width=.075\textwidth]{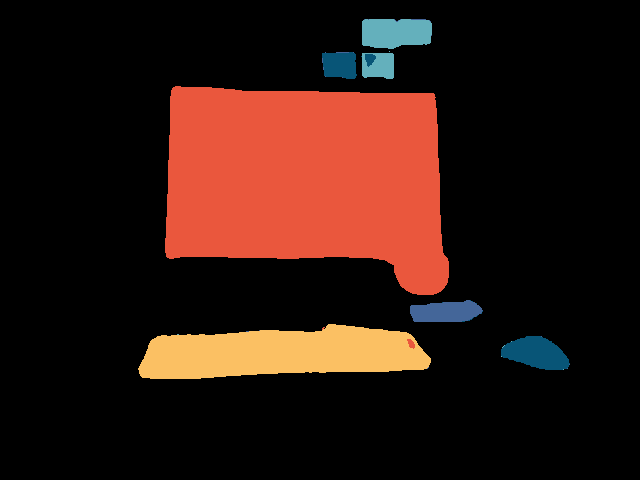}
\includegraphics[width=.075\textwidth]{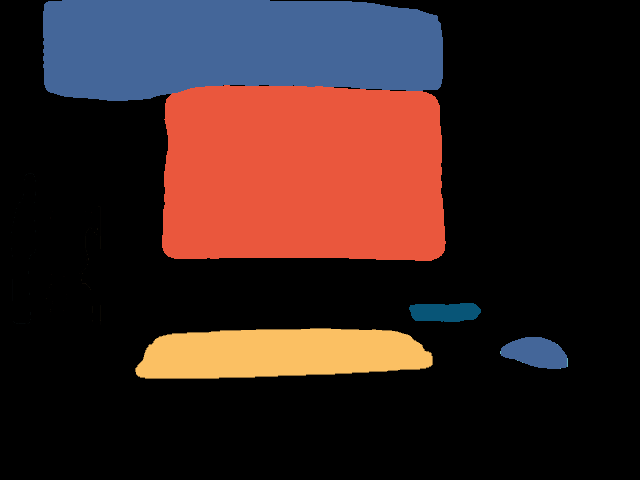}
\includegraphics[width=.075\textwidth]{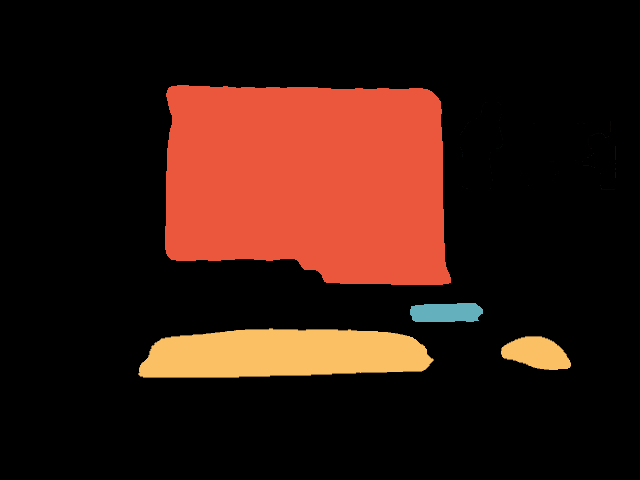}
\includegraphics[width=.075\textwidth]{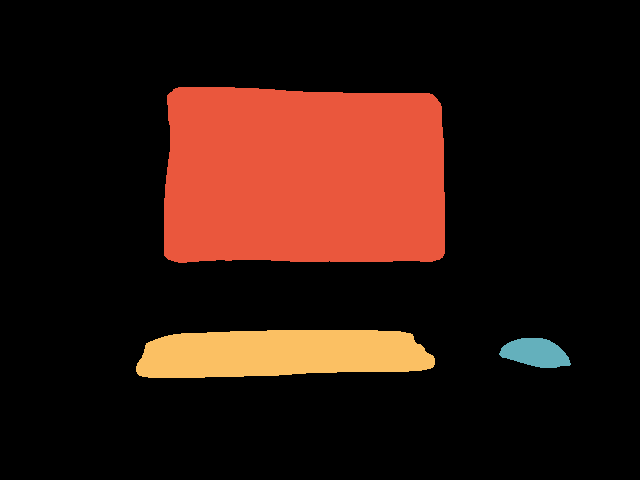}
\includegraphics[width=.075\textwidth]{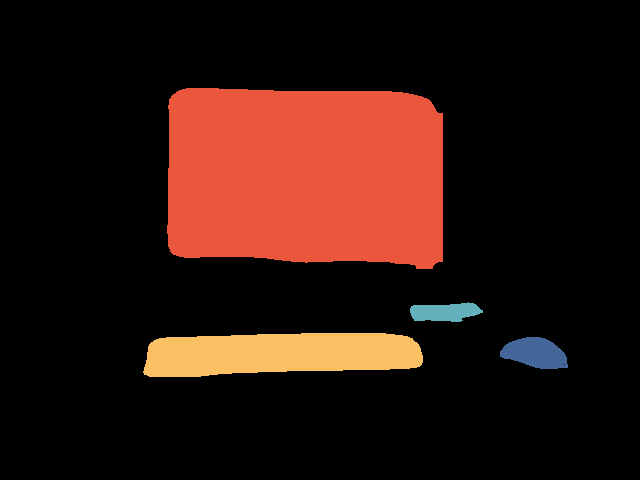}
\includegraphics[width=.075\textwidth]{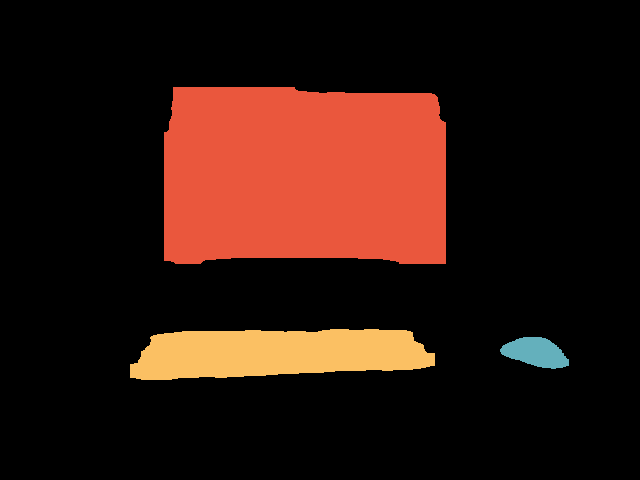}
\includegraphics[width=.075\textwidth]{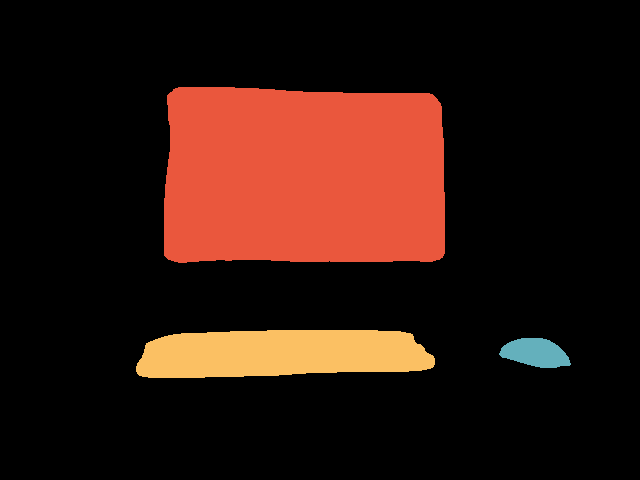}
\includegraphics[width=.075\textwidth]{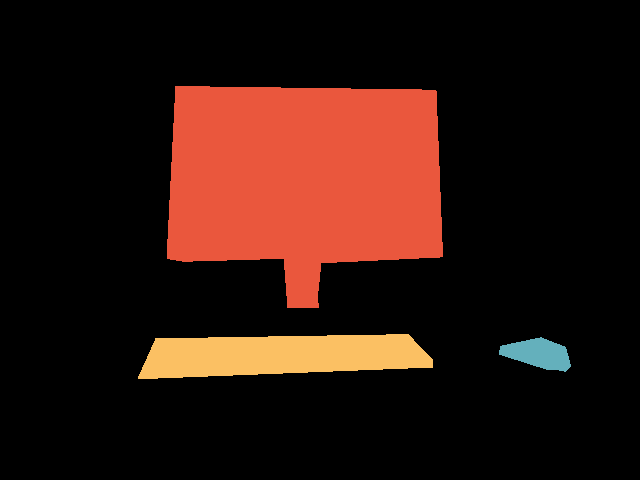}
\\
\vspace{0.1cm}
\includegraphics[width=.075\textwidth]{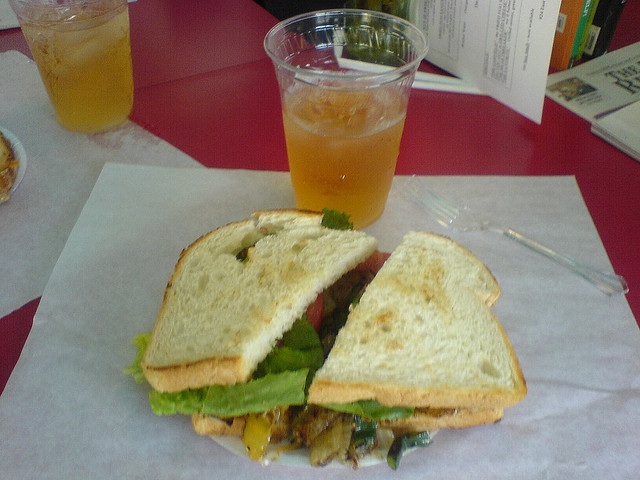}
\includegraphics[width=.075\textwidth]{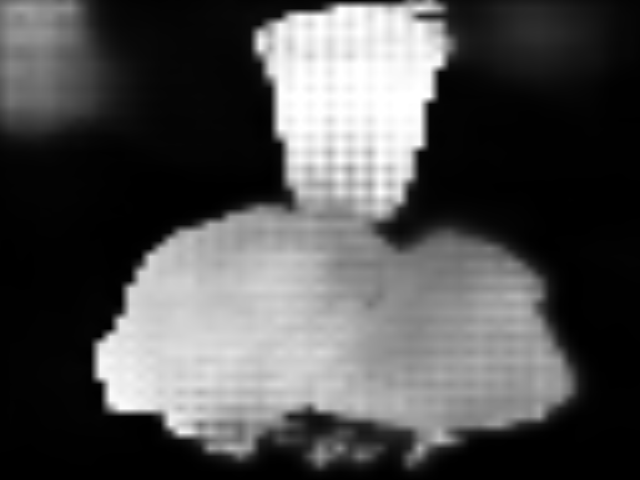}
\includegraphics[width=.075\textwidth]{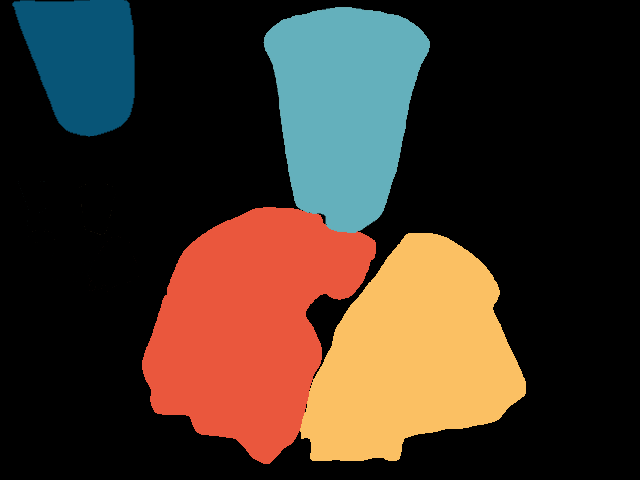}
\includegraphics[width=.075\textwidth]{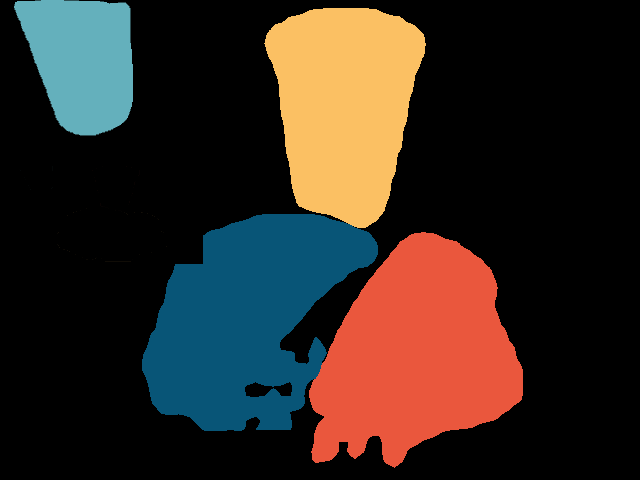}
\includegraphics[width=.075\textwidth]{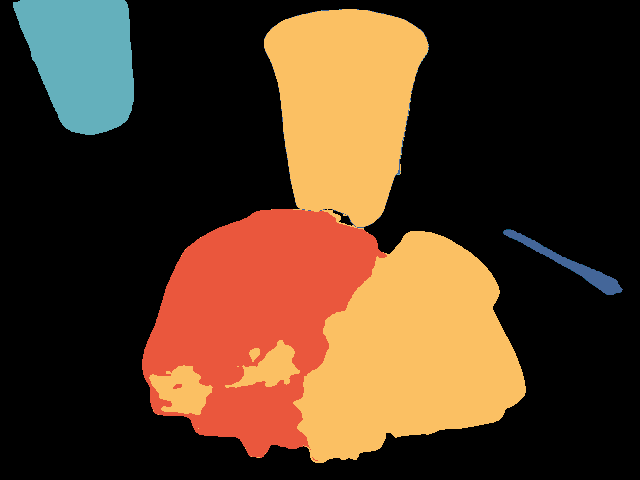}
\includegraphics[width=.075\textwidth]{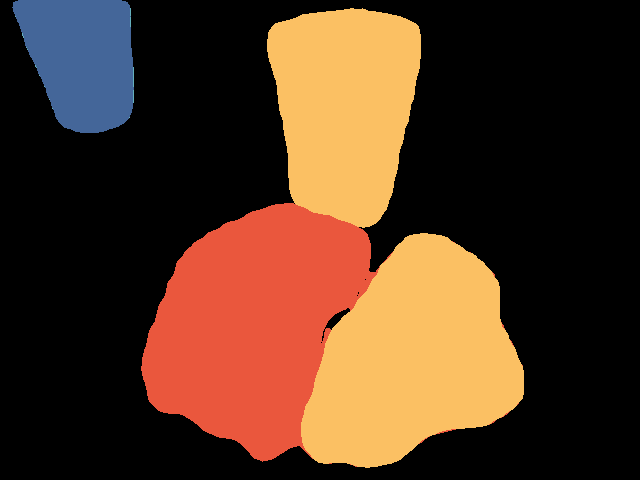}
\includegraphics[width=.075\textwidth]{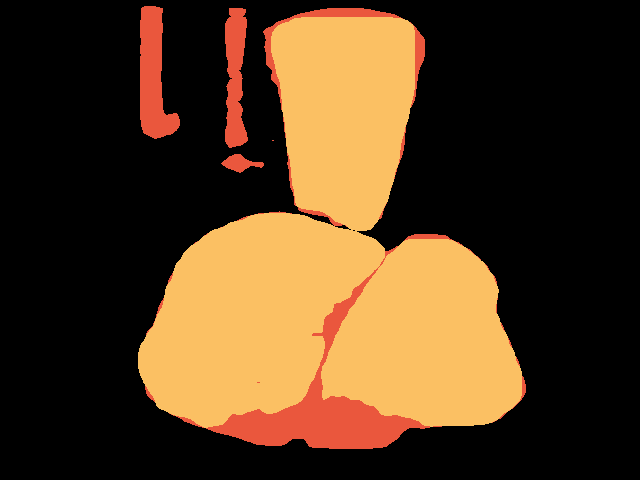}
\includegraphics[width=.075\textwidth]{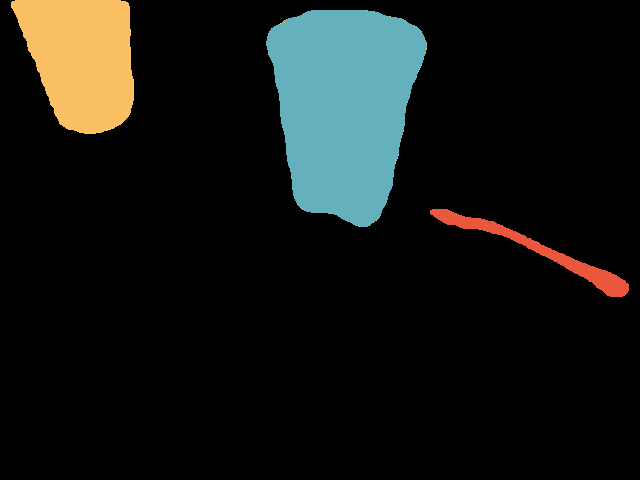}
\includegraphics[width=.075\textwidth]{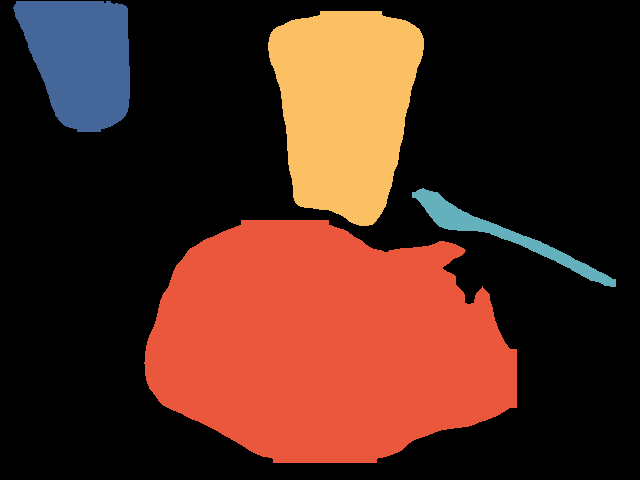}
\includegraphics[width=.075\textwidth]{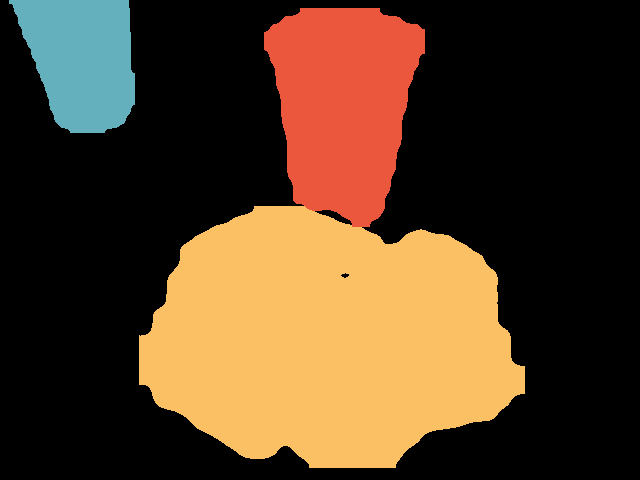}
\includegraphics[width=.075\textwidth]{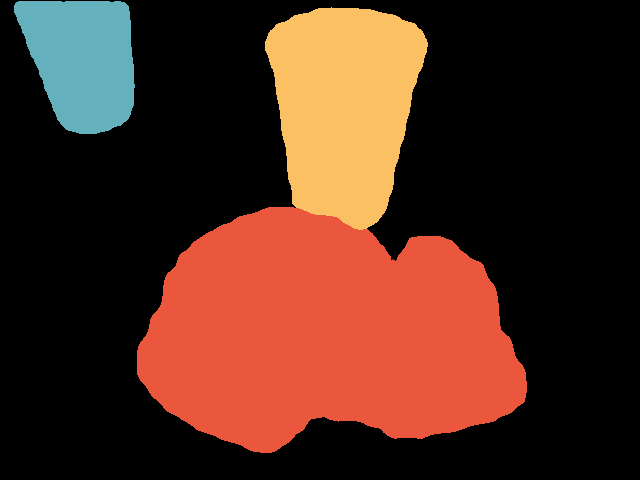}
\includegraphics[width=.075\textwidth]{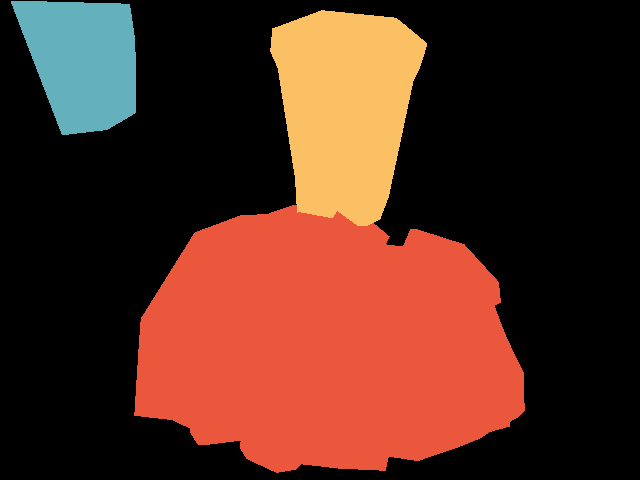}
\\
\scriptsize{\bf\underline{(3) Examples of ranking salient objects based on their semantics and positional contexts.}}
\\
\vspace{0.08cm}
\scriptsize{\underline{Row 7: two sinks at different distances and a small liquid soap in the middle, and Row 8: a nearby cat and a distant bird.}}
\\
\vspace{0.07cm}
\includegraphics[width=.075\textwidth]{quality_res/5/COCO_val2014_000000167443.jpg}
\includegraphics[width=.075\textwidth]{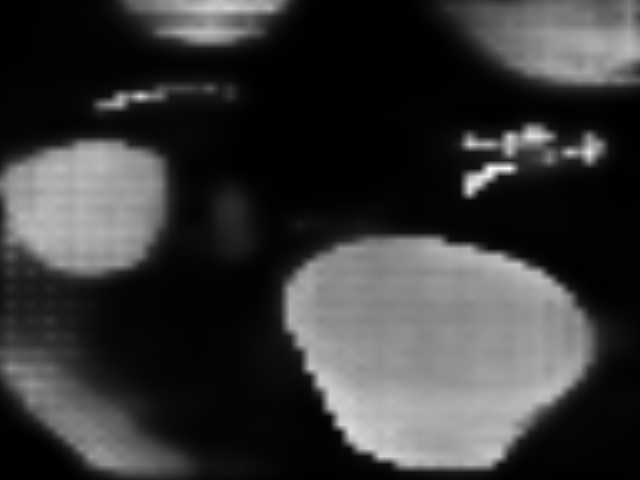}
\includegraphics[width=.075\textwidth]{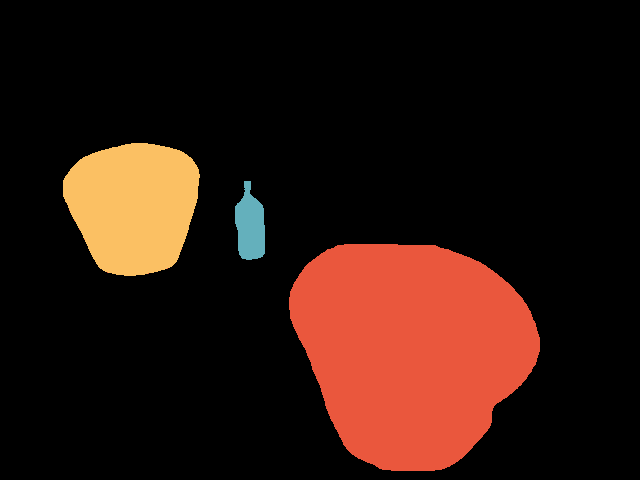}
\includegraphics[width=.075\textwidth]{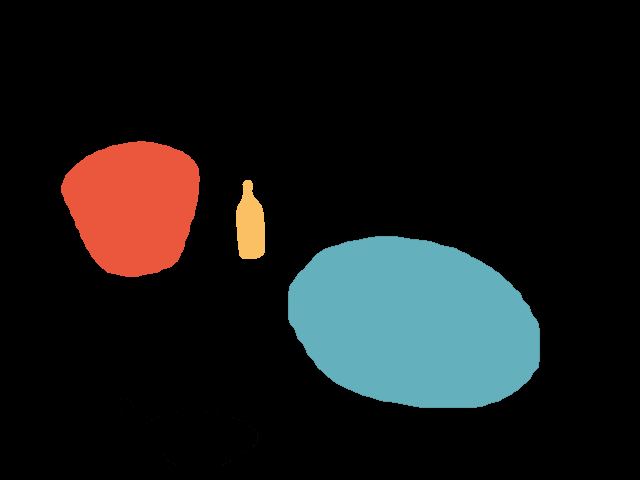}
\includegraphics[width=.075\textwidth]{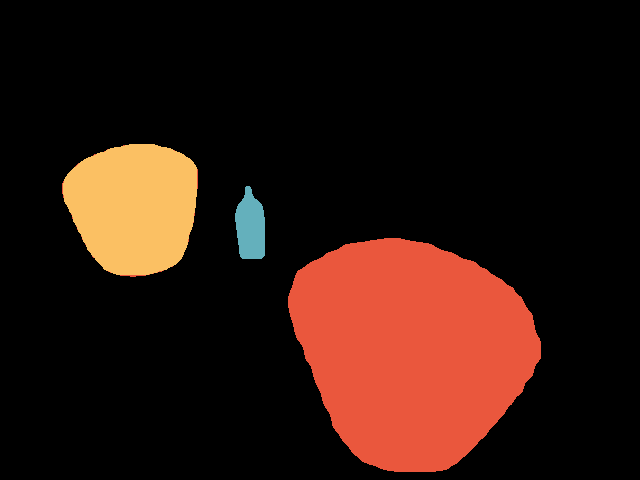}
\includegraphics[width=.075\textwidth]{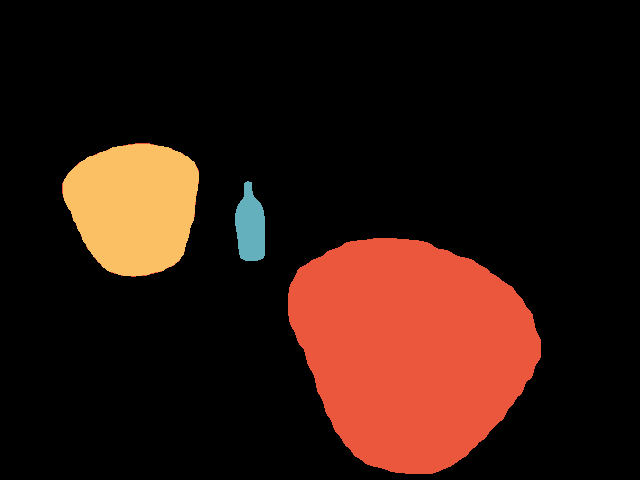}
\includegraphics[width=.075\textwidth]{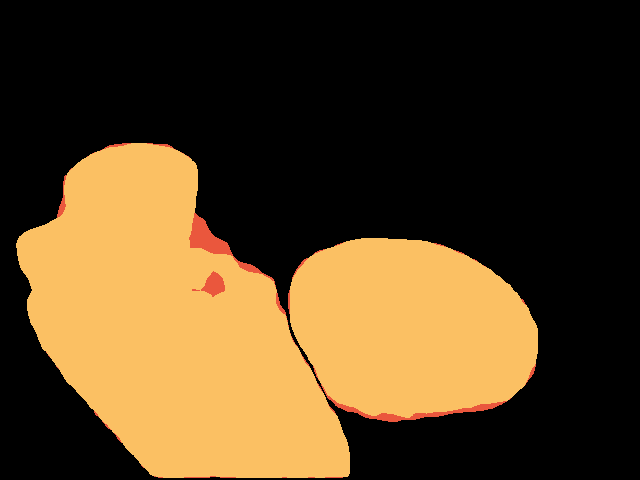}
\includegraphics[width=.075\textwidth]{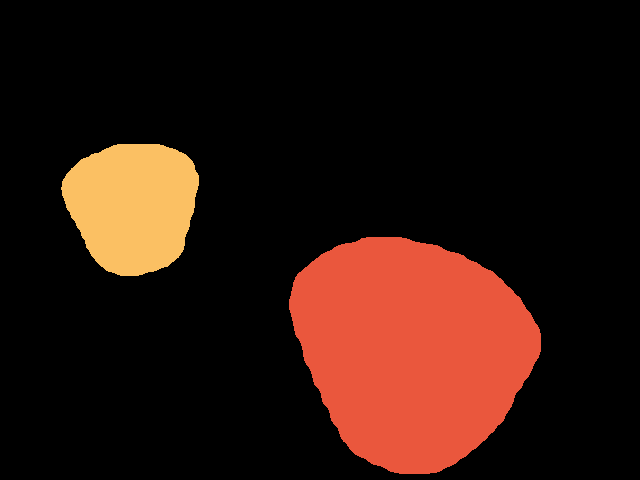}
\includegraphics[width=.075\textwidth]{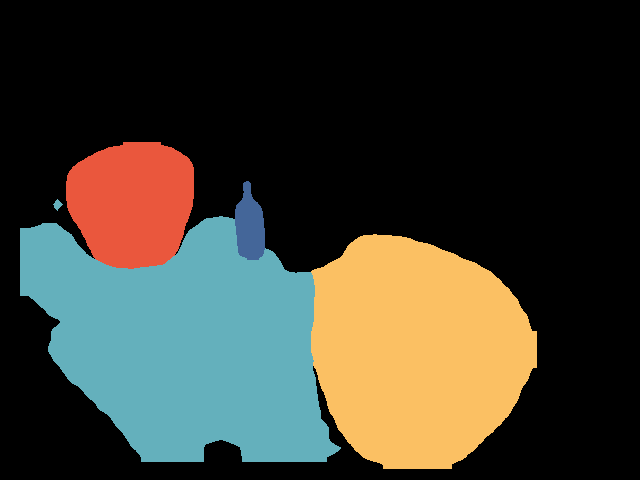}
\includegraphics[width=.075\textwidth]{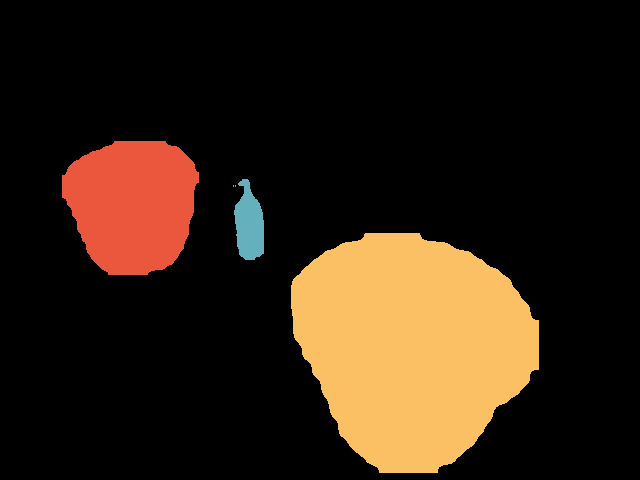}
\includegraphics[width=.075\textwidth]{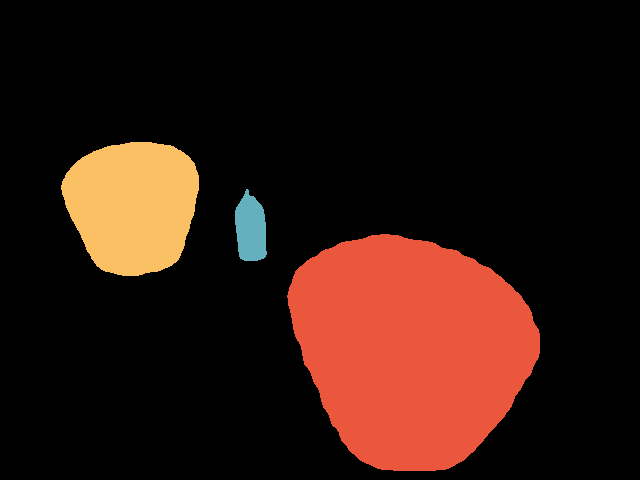}
\includegraphics[width=.075\textwidth]{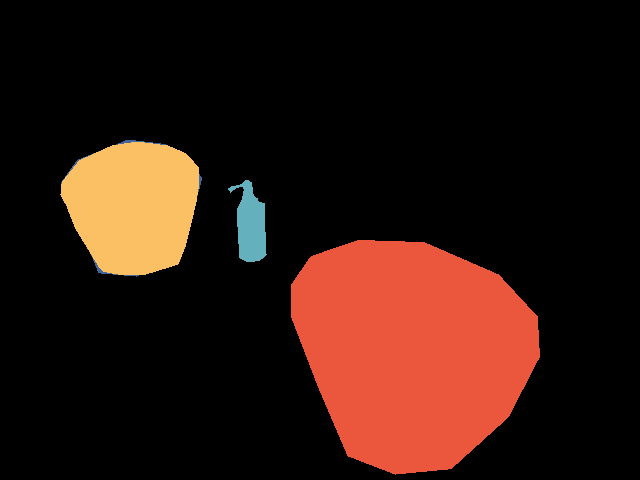}
\\
\vspace{0.07cm}
\includegraphics[width=.075\textwidth]{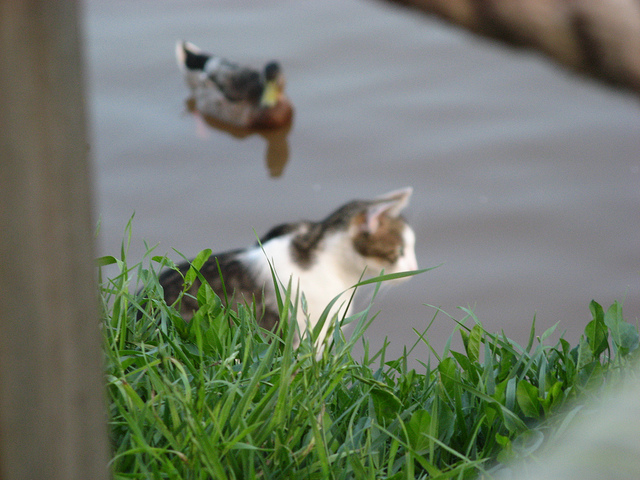}
\includegraphics[width=.075\textwidth]{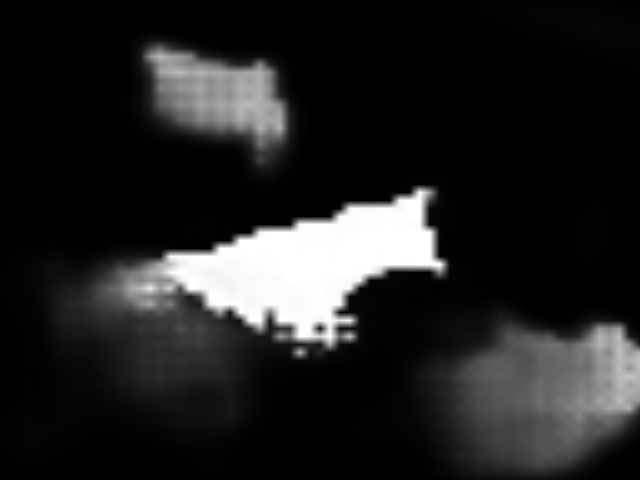}
\includegraphics[width=.075\textwidth]{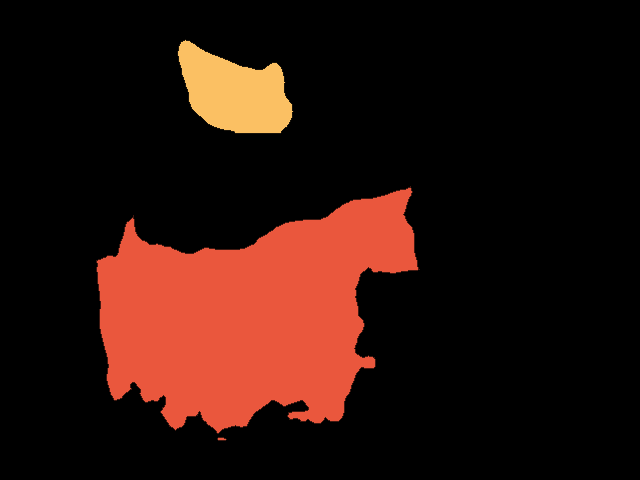}
\includegraphics[width=.075\textwidth]{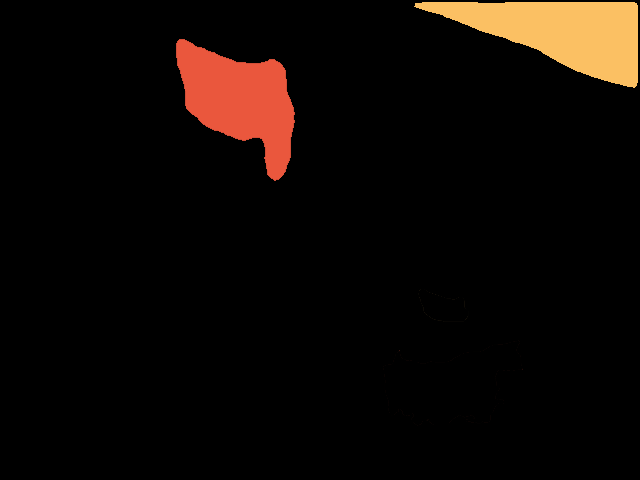}
\includegraphics[width=.075\textwidth]{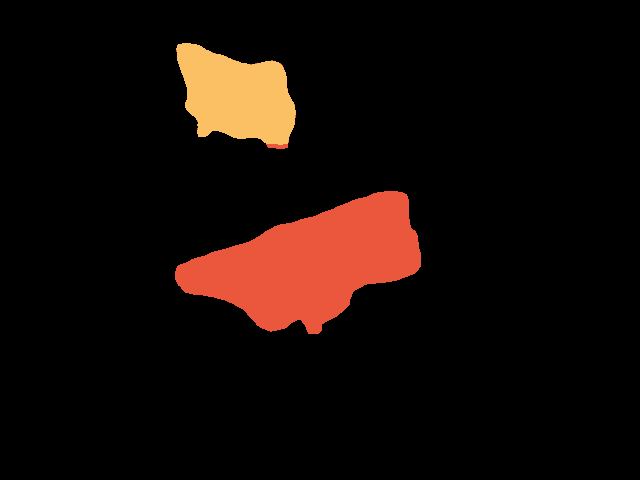}
\includegraphics[width=.075\textwidth]{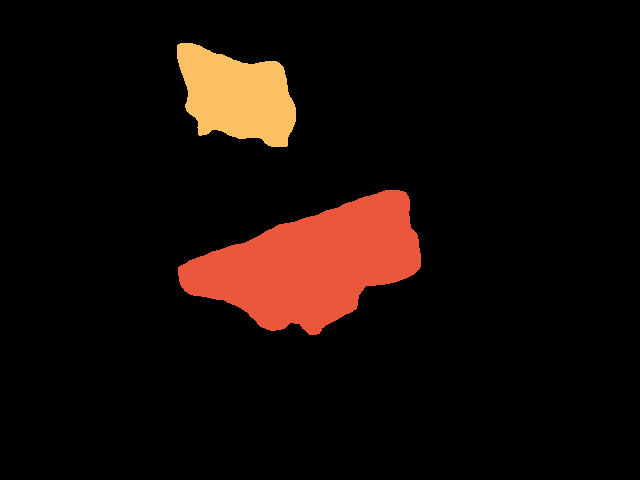}
\includegraphics[width=.075\textwidth]{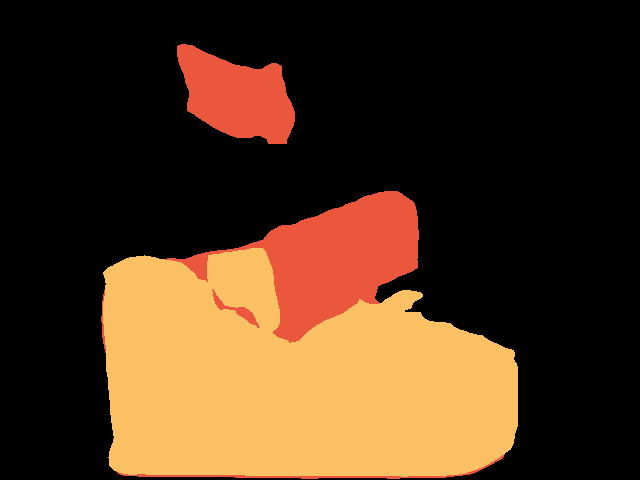}
\includegraphics[width=.075\textwidth]{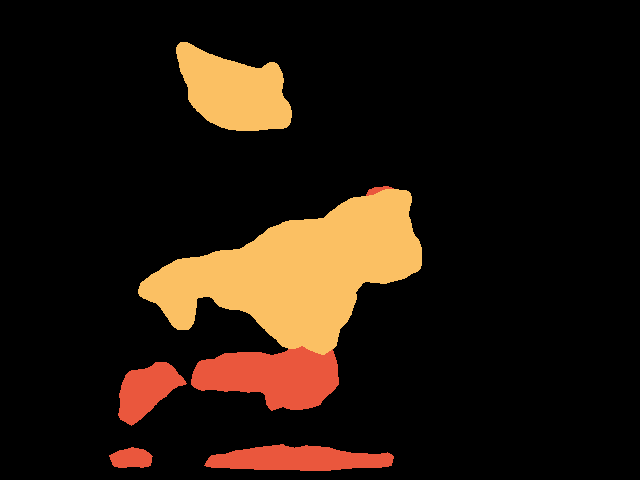}
\includegraphics[width=.075\textwidth]{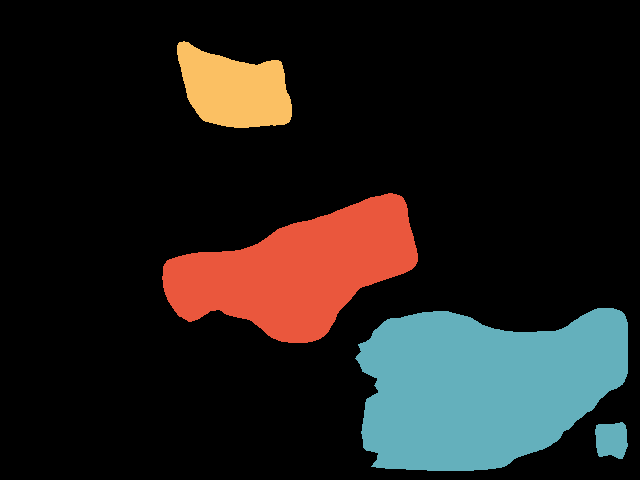}
\includegraphics[width=.075\textwidth]{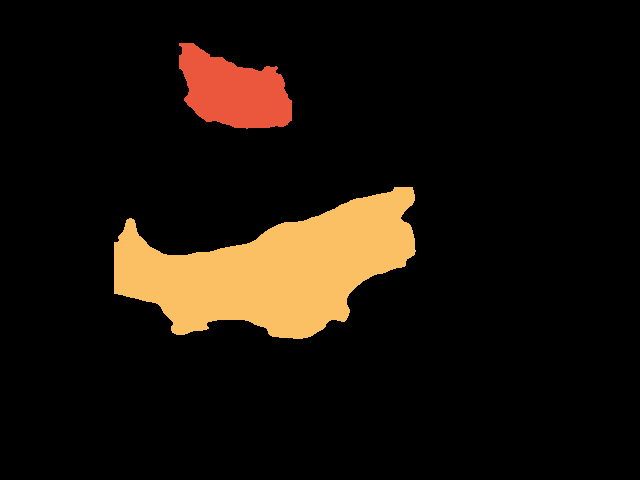}
\includegraphics[width=.075\textwidth]{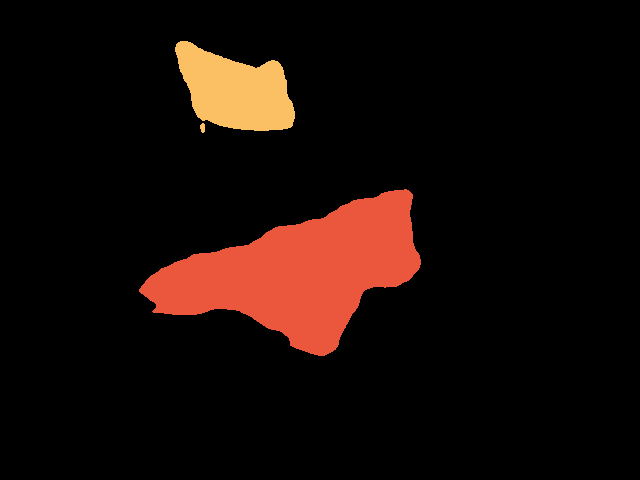}
\includegraphics[width=.075\textwidth]{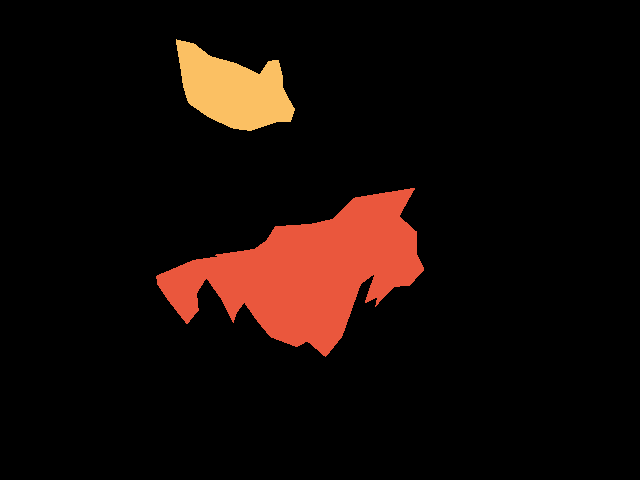}
\\
\scriptsize{\bf\underline{(4) Examples of ranking salient objects based on their semantics and low-level stimulus.}}
\\
\vspace{0.08cm}
\scriptsize{\underline{Row 9: a man holding a colorful kite, and Row 10: a woman, and a girl grabbing a colorful umbrella.}}
\\
\vspace{0.07cm}
\includegraphics[width=.075\textwidth]{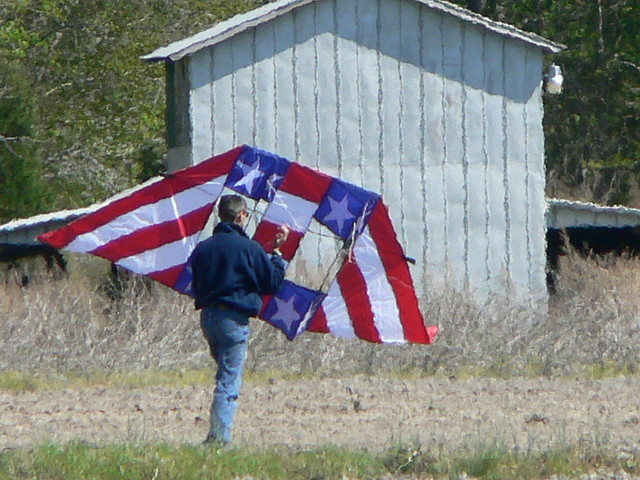}
\includegraphics[width=.075\textwidth]{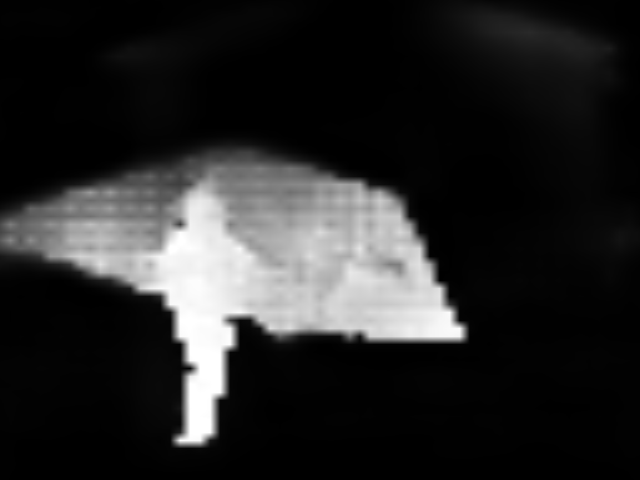}
\includegraphics[width=.075\textwidth]{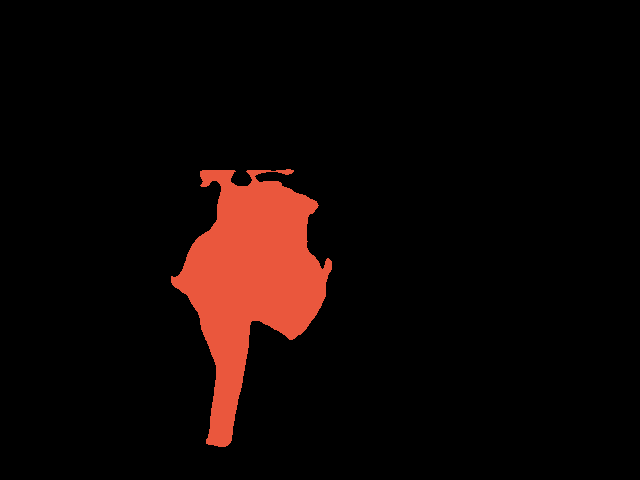}
\includegraphics[width=.075\textwidth]{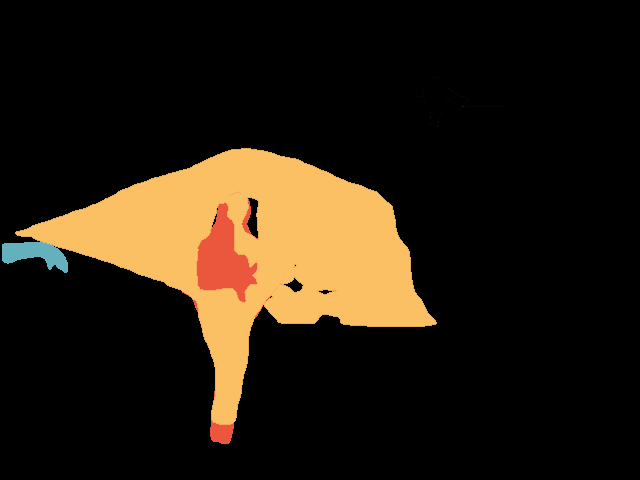}
\includegraphics[width=.075\textwidth]{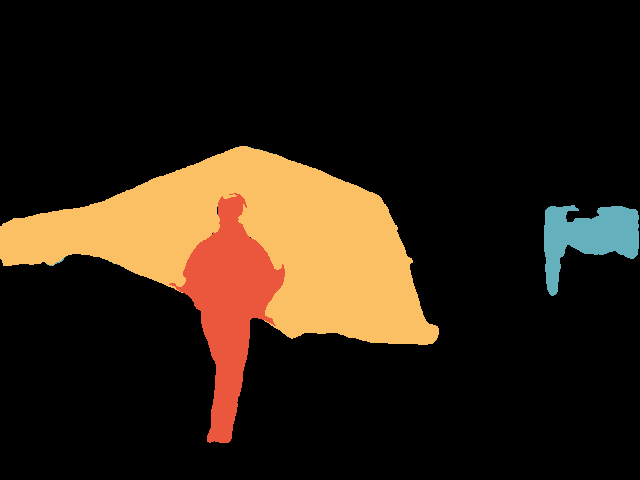}
\includegraphics[width=.075\textwidth]{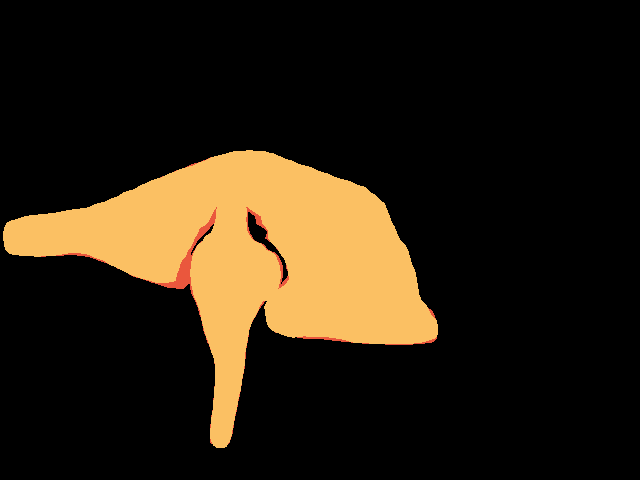}
\includegraphics[width=.075\textwidth]{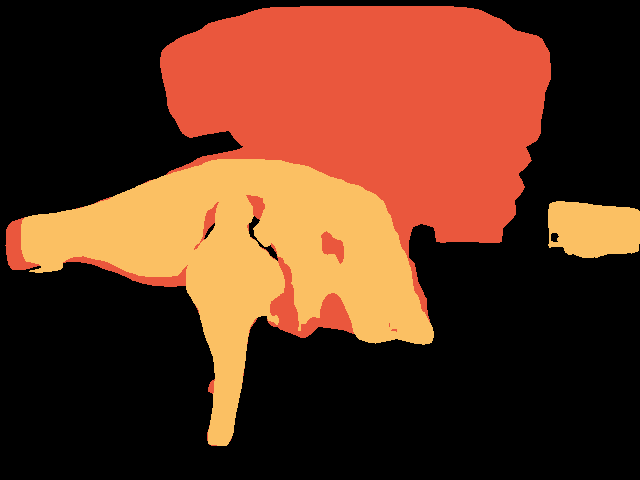}
\includegraphics[width=.075\textwidth]{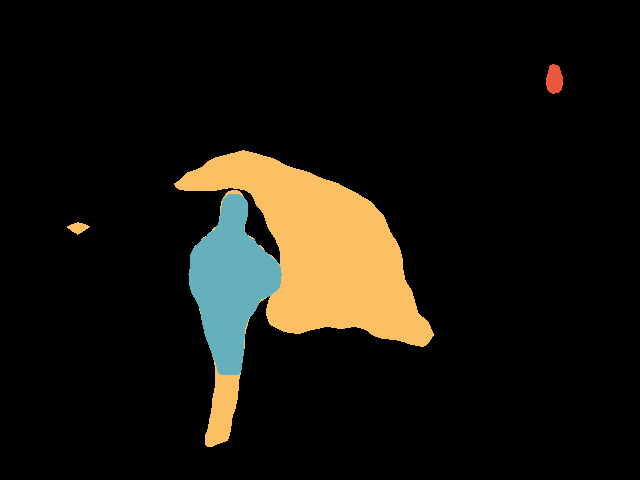}
\includegraphics[width=.075\textwidth]{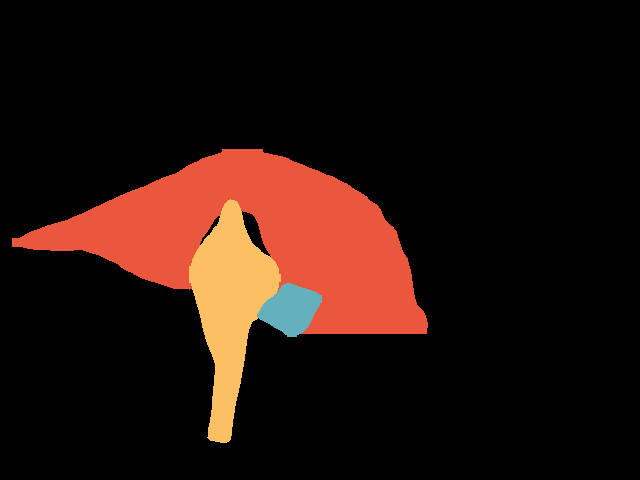}
\includegraphics[width=.075\textwidth]{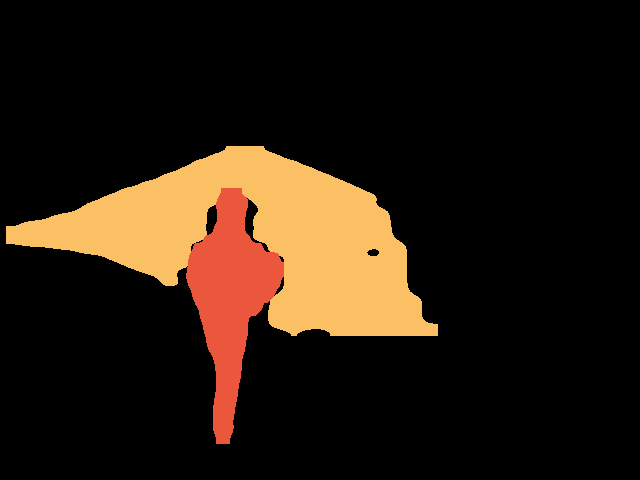}
\includegraphics[width=.075\textwidth]{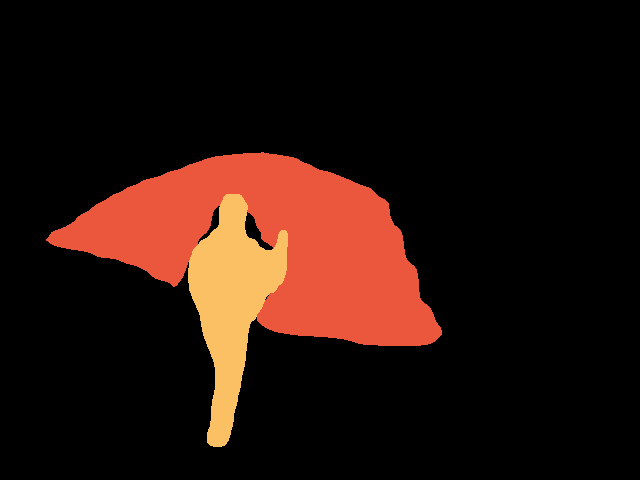}
\includegraphics[width=.075\textwidth]{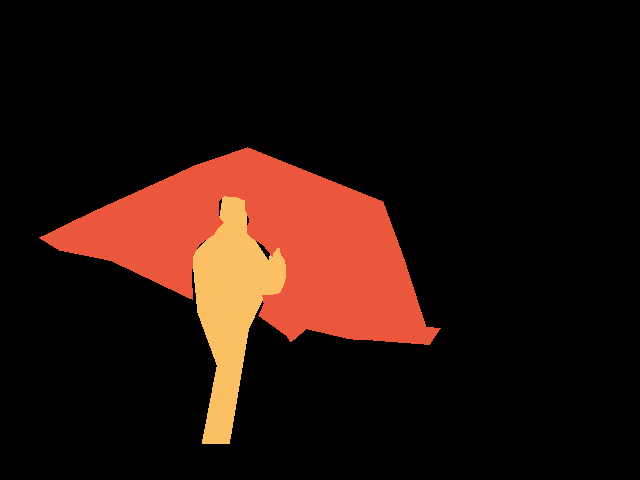}
\\
\vspace{0.07cm}
\includegraphics[width=.075\textwidth]{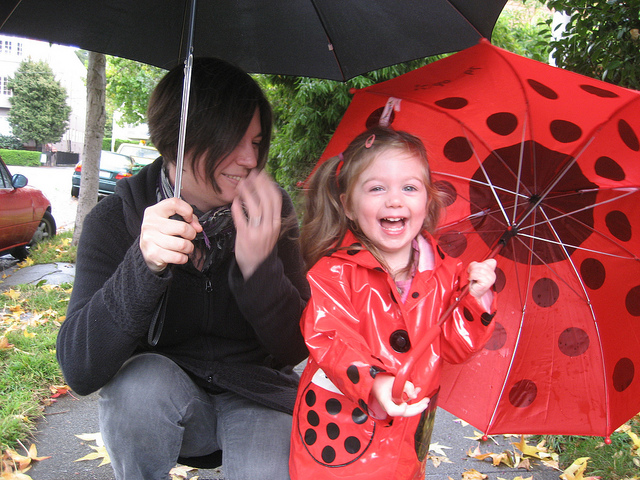}
\includegraphics[width=.075\textwidth]{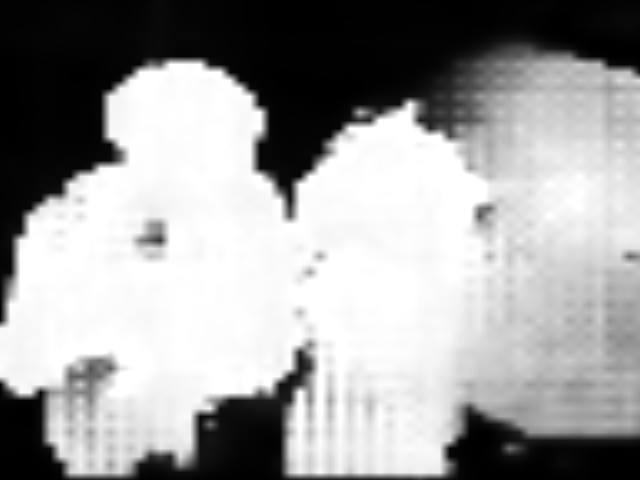}
\includegraphics[width=.075\textwidth]{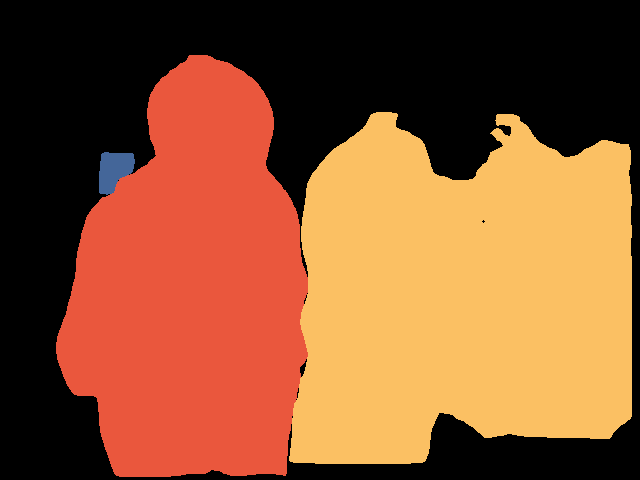}
\includegraphics[width=.075\textwidth]{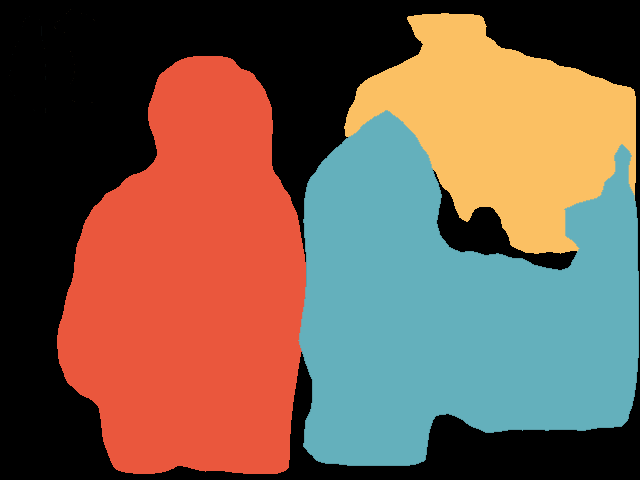}
\includegraphics[width=.075\textwidth]{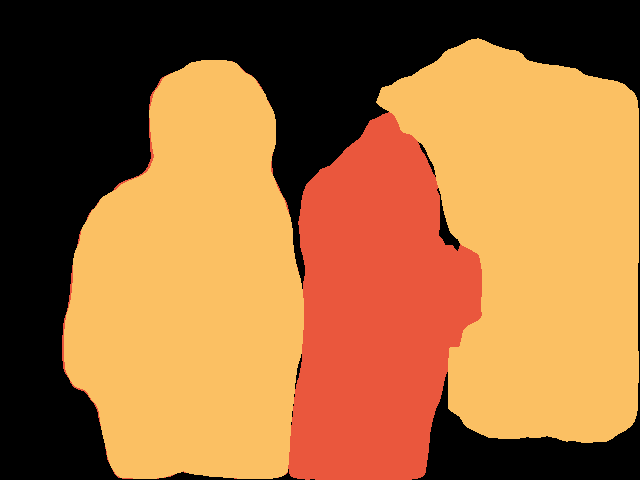}
\includegraphics[width=.075\textwidth]{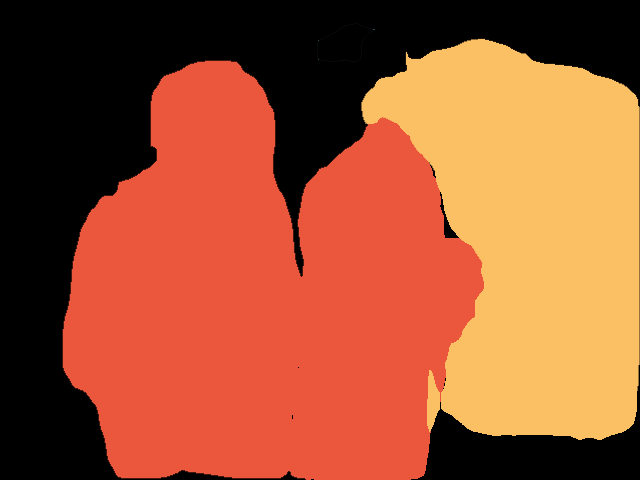}
\includegraphics[width=.075\textwidth]{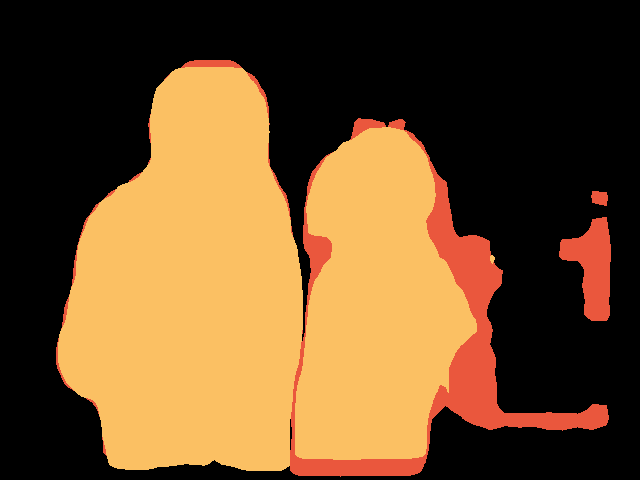}
\includegraphics[width=.075\textwidth]{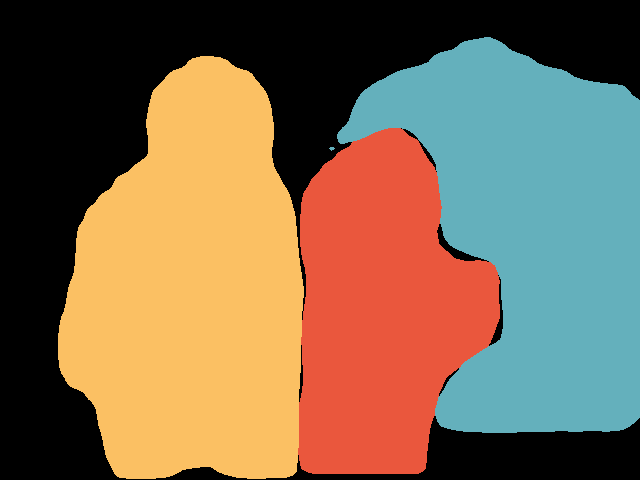}
\includegraphics[width=.075\textwidth]{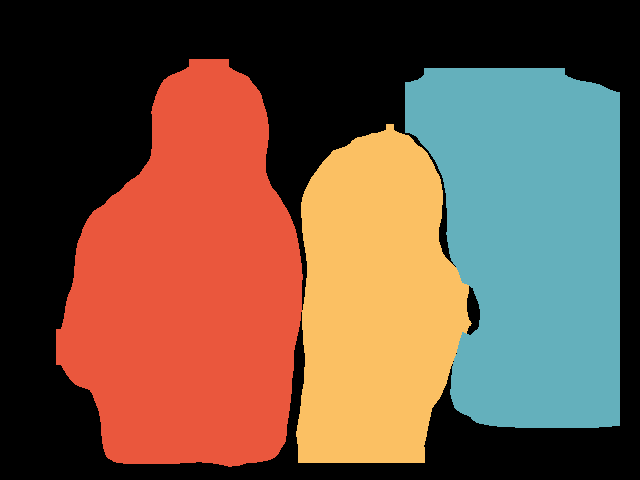}
\includegraphics[width=.075\textwidth]{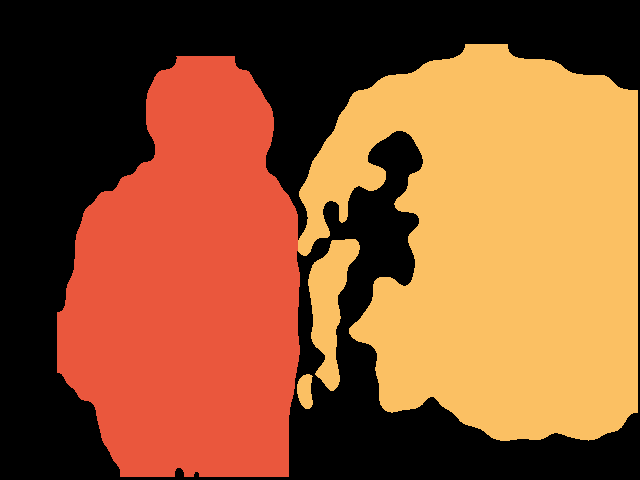}
\includegraphics[width=.075\textwidth]{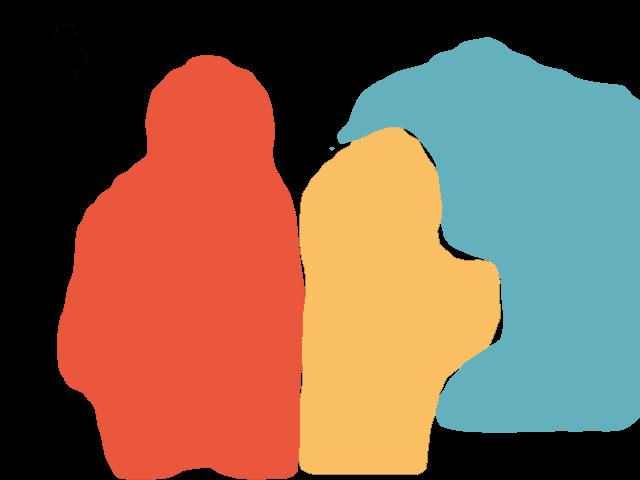}
\includegraphics[width=.075\textwidth]{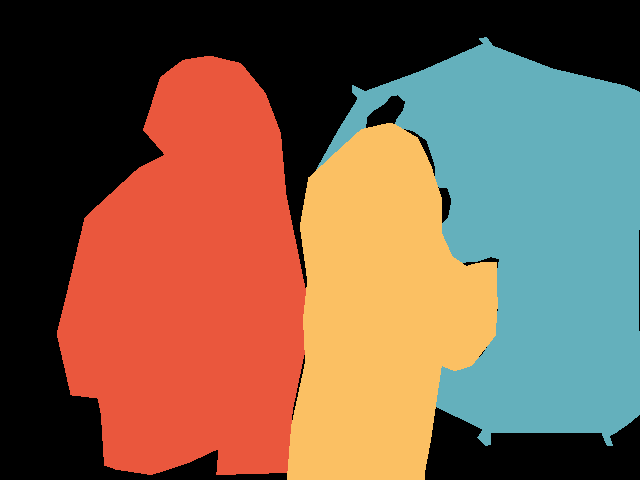}
\\
\vspace{0.20mm}
\centering
\includegraphics[width=0.85\textwidth]{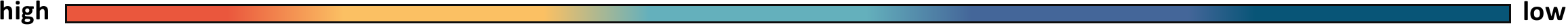}
\vspace{-0.15cm}
\caption{Qualitative comparison of our method with existing saliency ranking approaches and baseline methods. \tx{Above each image group, we summarize the scenarios of the images used for the test.}
}
\label{fig:quality_res}
\end{figure*}

\noindent{\bf Qualitative Comparisons.} We further qualitatively evaluate our method, as shown in Figure \ref{fig:quality_res}. Due to limited space, we only compare with the best performing methods in Table~\ref{tab:main}. 
We can see that existing methods commonly suffer from   detecting non-salient objects, inferring incorrect saliency ranks among objects, and under-/over-detection of objects.
The visual comparison show that our method is able to individualize salient objects and infer their saliency ranks accurately for diverse scenes.

\subsection{Module Analysis}

\noindent{\bf Ablation Study of Our Model.} We begin by studying the effectiveness of the proposed SOS and OCOR modules. Table \ref{tab:ablation} shows the results. By removing each or both of them from the network, we can see that the performance drops significantly. This shows both SOS and OCOR modules are important to the saliency ranking performance.

\begin{table}[!h]\small
\caption{Ablation study of the SOS and OCOR modules.}
\vspace{-5mm}
\begin{center}
\begin{tabular}{c c c c}
\toprule[1.5pt]
\textbf{Method} & \textbf{SA-SOR $\uparrow$} & \textbf{SOR$\uparrow$} & \textbf{MAE$\downarrow$} \\
\midrule[1.0pt]
\makecell[c]{w/o SOS, OCOR} & 0.715 & 0.837 & 0.095 \\
w/o SOS & 0.729 & 0.885 & 0.084 \\
w/o OCOR & 0.722 & 0.870 & 0.090 \\
Ours  & {\bf 0.738} & {\bf 0.904} & {\bf 0.078} \\
\bottomrule[1.5pt]
\end{tabular}
\end{center}
\label{tab:ablation}
\vspace{-5mm}
\end{table}

\noindent{\bf Internal Analysis of the SOS Module.} We study how Global Covariance Pooling (GCP) and Dynamic Rectifying (DR) functions in the SOS module influence the saliency ranking performance. Table \ref{tab:sos} shows the results, from which we can see that GCP and DR perform much more better when they work together than individually.

\begin{table}[t]\small
\caption{Internal analysis of the SOS module.}
\vspace{-5mm}
\begin{center}
\begin{tabular}{c c c c}
\toprule[1.5pt]
\textbf{Method} & \textbf{SA-SOR $\uparrow$} & \textbf{SOR$\uparrow$} & \textbf{MAE$\downarrow$} \\
\midrule[1.0pt]
Baseline & 0.715 & 0.837 & 0.095 \\
Baseline + GCP & 0.715 & 0.840 & 0.092 \\
Baseline + DR & 0.719 & 0.857 & 0.094 \\
Baseline + SOS & {\bf 0.722} & {\bf 0.870} & {\bf 0.090} \\
\bottomrule[1.5pt]
\end{tabular}
\end{center}
\label{tab:sos}
\vspace{-5mm}
\end{table}

\vspace{1mm}
\noindent{\bf Alternatives of the OCOR Module.} Finally, we exploit four alternative strategies to build bi-directional object-context relations. {\bf S1}: Given two object features and the context features, we concatenate them together and feed them to two conv. layers to learn their context relation; {\bf S2}: Based on {S1}, we apply channel-wise attention \cite{woo2018cbam} on the fused features; {\bf S3}: Based on {S1}, we apply spatial-wise attention \cite{woo2018cbam} on the fused features; {\bf S4}: Based on {S1}, we apply both channel-wise and spatial-wise attentions \cite{woo2018cbam} on the fused features.
Table~\ref{tab:ocor} shows that our OCOR module outperforms all these baselines, verifying that our OCOR module can capture region-level context relating to an object and build bidirectional object-context relations well.

\begin{table}[t]\small
\caption{Comparison of different strategies for modeling object-context-object relation.}
\vspace{-5mm}
\begin{center}
\begin{tabular}{c c c c}
\toprule[1.5pt]
\textbf{Method} & \textbf{SA-SOR $\uparrow$} & \textbf{SOR$\uparrow$} & \textbf{MAE$\downarrow$} \\
\midrule[1.0pt]
Baseline & 0.715 & 0.837 & 0.095 \\
Baseline + S1 & 0.718 & 0.842 & 0.092 \\
Baseline + S2 & 0.719 & 0.845 & 0.093 \\
Baseline + S3 & 0.722 & 0.853 & 0.090 \\
Baseline + S4 & 0.722 & 0.857 & 0.088 \\
Baseline + OCOR & {\bf 0.729} & {\bf 0.885} & {\bf 0.084} \\
\bottomrule[1.5pt]
\end{tabular}
\end{center}
\label{tab:ocor}
\vspace{-5mm}
\end{table}


\begin{figure}[t]
\centering
\begin{minipage}[t]{0.14\textwidth}
\centering
{{\small{Input image}}}
\end{minipage}
\begin{minipage}[t]{0.14\textwidth}
\centering
{{\small{Our prediction}}}
\end{minipage}
\begin{minipage}[t]{0.14\textwidth}
\centering
{{\small{Ground truth}}}
\end{minipage}
\\
    \centering
    \includegraphics[width=.14\textwidth]{limitation/COCO_val2014_000000058886.jpg}
    \centering
    \includegraphics[width=.14\textwidth]{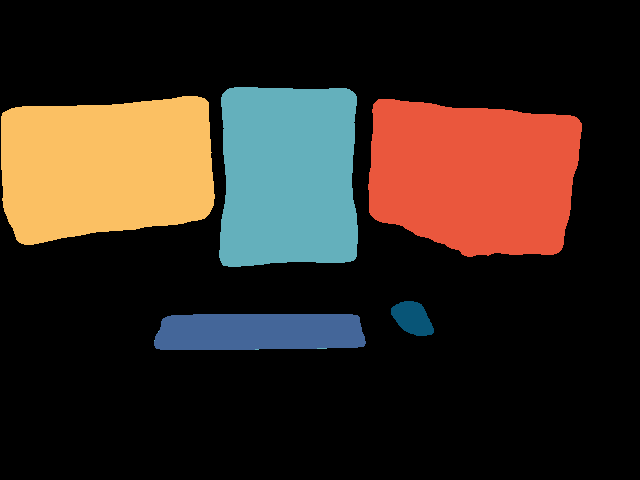}
    \centering
    \includegraphics[width=.14\textwidth]{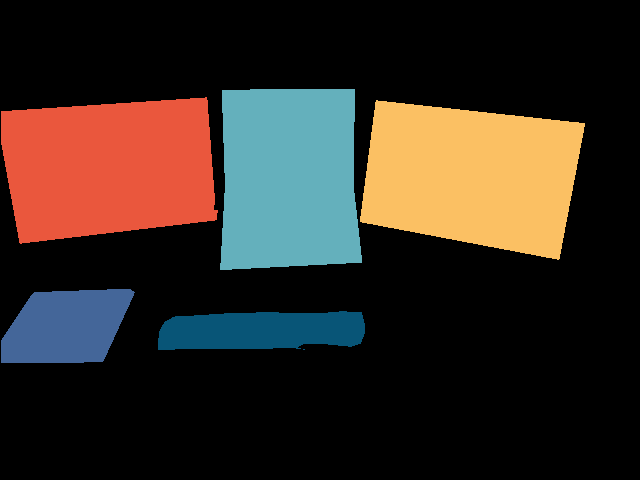}
    \\
    \vspace{0.07cm}
    \centering
    \includegraphics[width=.14\textwidth]{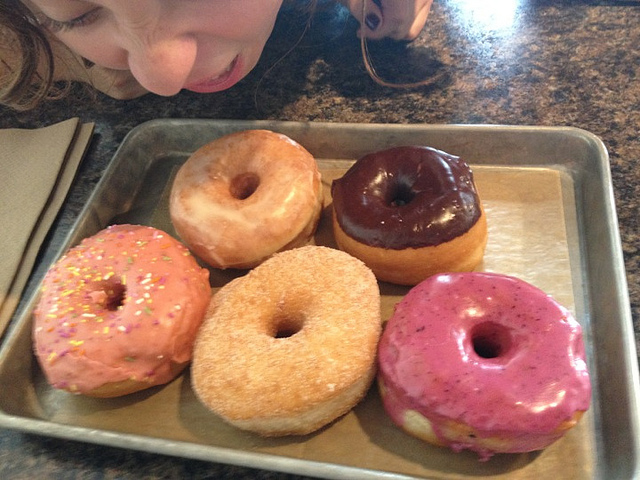}
    \includegraphics[width=.14\textwidth]{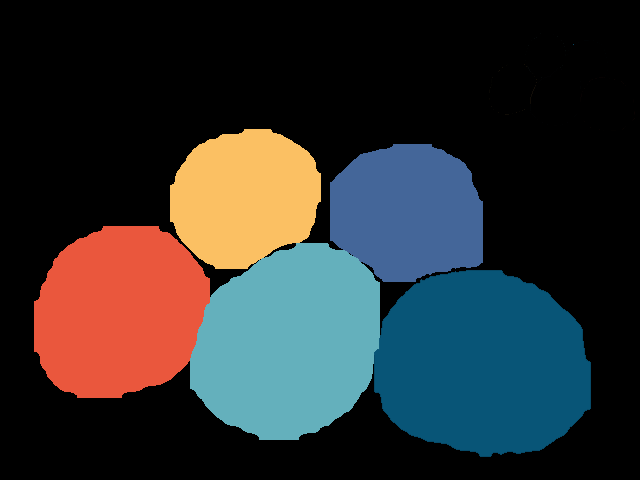}
    \includegraphics[width=.14\textwidth]{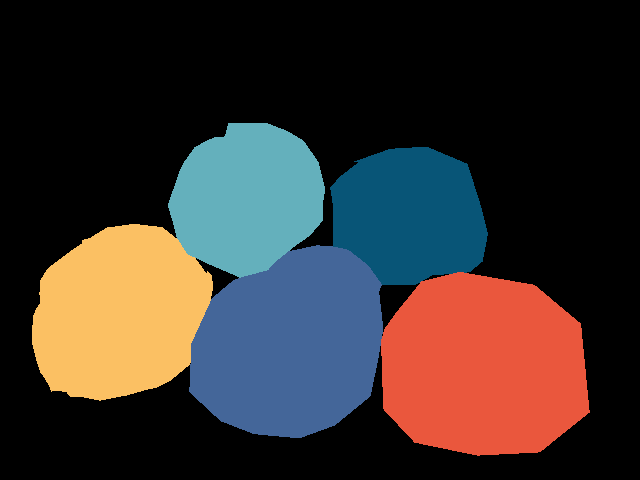}
    \caption{Our method may fail to infer accurate saliency ranks for objects with a similar functionality and interaction with the global contexts.}
    \label{fig:limitation}
\vspace{-5mm}
\end{figure}

\section{Conclusion}
In this paper, we have studied the saliency ranking problem. 
We observe that the human visual system leverage both spatial and object-based attentions for handling the visual inputs. We therefore propose a bi-directional Object-Context Prioritization Learning approach for saliency ranking.
A selective object saliency module is proposed to model object-based attention via capturing and enhancing object semantic representation. 
An object-context-object relation module is proposed to model spatial attention through studying how an object with its regional context would interact with other objects with their regional contexts.
Extensive experiments have verified the effectiveness of
our method against state-of-the-art methods.

Our work does have a limitation. When objects with the same functionality appear in a scene with a narrow context, our model may fail to infer correct saliency ranking. 
Figure~\ref{fig:limitation} shows two examples. In the first row, the left and right screens interact with the global context symmetrically. In the second row, these donuts are all visually attractive. We suspect that in these scenarios, even humans may not have a consistent ranking.
In the future, we would like to delve deeper into the human visual system for more cues to help with the saliency ranking task.

%

\noindent\ttxx{{\bf Acknowledgements:} This work was partly supported by the National Natural Science Foundation of China under Grants 61972067/U19B2039, the Innovation Technology Funding of Dalian (2020JJ26GX036), a General Research Fund from RGC of Hong Kong (RGC Ref.: 11205620), and a Strategic Research Grant from City University of Hong Kong (Ref.: 7005674).}

{\small
\bibliographystyle{ieee_fullname}
\bibliography{egbib}
}

\end{document}